\pdfoutput=1

\documentclass{article}
\usepackage{colm2024_conference}

\usepackage{subcaption}
\usepackage{url}
\usepackage{booktabs}
\usepackage{hyperref}
\usepackage{graphicx}
\usepackage{microtype}
\usepackage{multicol}
\usepackage{xcolor}   
\usepackage{multirow}
\usepackage{lipsum}
\usepackage{enumitem}
\usepackage{pifont}
\usepackage{colortbl}
\usepackage{tabularx}
\usepackage{caption} 
\usepackage{float}
\usepackage{amsmath,amsfonts,amssymb,bbm}

\newcommand*\samethanks[1][\value{footnote}]{\footnotemark[#1]}

\title{QUADS: Stabilizing NVFP4 Reinforcement Learning for MoE via QUantization-error Alignment across Dual Sides}

\author{
\textbf{Zhengyang Zhuge}\thanks{Equal contribution.} \quad
\textbf{Hao Yu}\samethanks \quad
\textbf{Xin Wang}\samethanks \quad
\textbf{Zheng Li} \quad
\textbf{Yizhong Cao} \quad
\textbf{Dayiheng Liu} \quad 
\textbf{Jianwei Zhang}\thanks{Corresponding author.} \quad
\\
\vspace{1.5mm}
Qwen Team, Alibaba Inc.
}

\begin{document}
\maketitle

\begin{abstract}
    Rollout generation is a major bottleneck in Reinforcement Learning (RL) for Mixture-of-Experts (MoE) Large Language Models, motivating low-precision rollout acceleration such as FP8.
    As an emerging low-precision format, NVFP4 combines fine-grained scaling for accuracy preservation with native W4A4 FP4 GEMMs for higher throughput than FP8.
    However, we find that directly applying NVFP4 to MoE RL rollout is impractical.
    NVFP4 rollout with BF16 training collapses after roughly 150 steps, accompanied by rapidly growing rollout--trainer log-probability gaps.
    Through training--inference error analysis and controlled ablations, we identify activation error, rather than weight error, as the dominant source of FP4 RL instability: weights can be synchronized and aligned by a shared quantization--dequantization path, whereas activations are recomputed online and error is amplified by the coarse E2M1 grid.
    Therefore, to stabilize NVFP4 RL for MoE, we propose \textbf{QU}antization-error \textbf{A}lignment across \textbf{D}ual \textbf{S}ides (\textbf{QUADS}).
    On the trainer side, we introduce Asymmetric Quantization-Aware Training fake-quantizing weights while keeping activations unquantized for better alignment.
    On the rollout side, Residual Activation Compensation corrects high-error activation channels while preserving native W4A4 GEMMs.
    In our MoE RL experiments on several benchmarks, QUADS achieves BF16-level accuracy, improves average pass@1 by 21.49 points over naive NVFP4 RL, and delivers $\sim$16\% higher rollout throughput than FP8.
    \end{abstract}

\section{Introduction}
\label{sec:intro}

Reinforcement learning (RL) has become a central paradigm for scaling the reasoning capabilities of Large Language Models (LLMs)~\citep{shao2024deepseekmath}.
Modern pipelines decouple autoregressive \emph{rollout} on an inference engine from gradient computation on a separate trainer, yet rollout generation still dominates end-to-end wall-clock time---often exceeding 70\% of each training step~\citep{li2026qurl,xi2026jetrl,qiu2026fp8rl}.
As models generate longer chain-of-thought traces, this imbalance intensifies, making rollout acceleration a first-order systems priority for policy-gradient methods such as GRPO~\citep{shao2024deepseekmath}.
The challenge is that rollout precision is not merely a serving detail: token-level log-probabilities produced by the rollout engine determine importance ratios in the trainer, so numerical bias can directly corrupt the policy gradient.

A common acceleration strategy is to run rollout inference at reduced precision.
At the FP8 tier, recent RL systems have shown that the resulting \emph{training--inference precision mismatch} can often be controlled with truncated importance sampling (TIS)~\citep{yao2025offpolicy}, adaptive clipping, or unified FP8 precision flows~\citep{li2026qurl,xi2026jetrl,qiu2026fp8rl,zhou2026ais}.
This makes FP8 a strong practical baseline: it accelerates rollout while keeping the log-probability gap small enough for stable optimization.

\begin{figure}[t]
    \centering
    \begin{minipage}{0.32\linewidth}
        \centering
        \includegraphics[width=\linewidth]{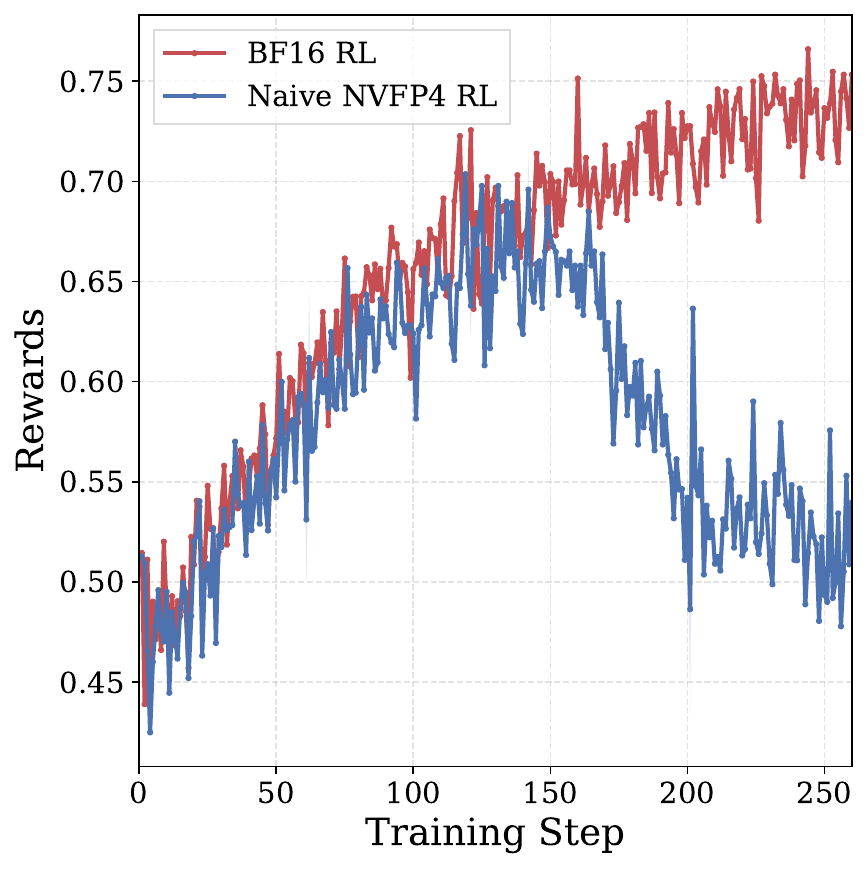}
        \centerline{\small (a) Training Reward}
    \end{minipage}\hfill
    \begin{minipage}{0.32\linewidth}
        \centering
        \includegraphics[width=\linewidth]{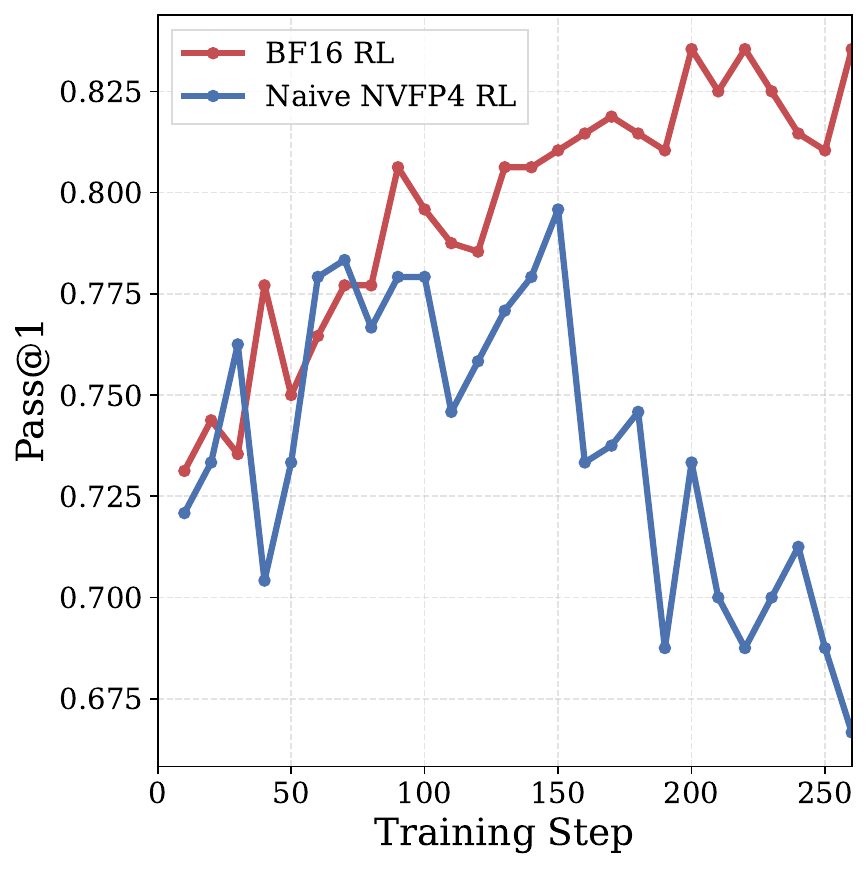}
        \centerline{\small (b) Test Score}
    \end{minipage}\hfill
    \begin{minipage}{0.32\linewidth}
        \centering
        \includegraphics[width=\linewidth]{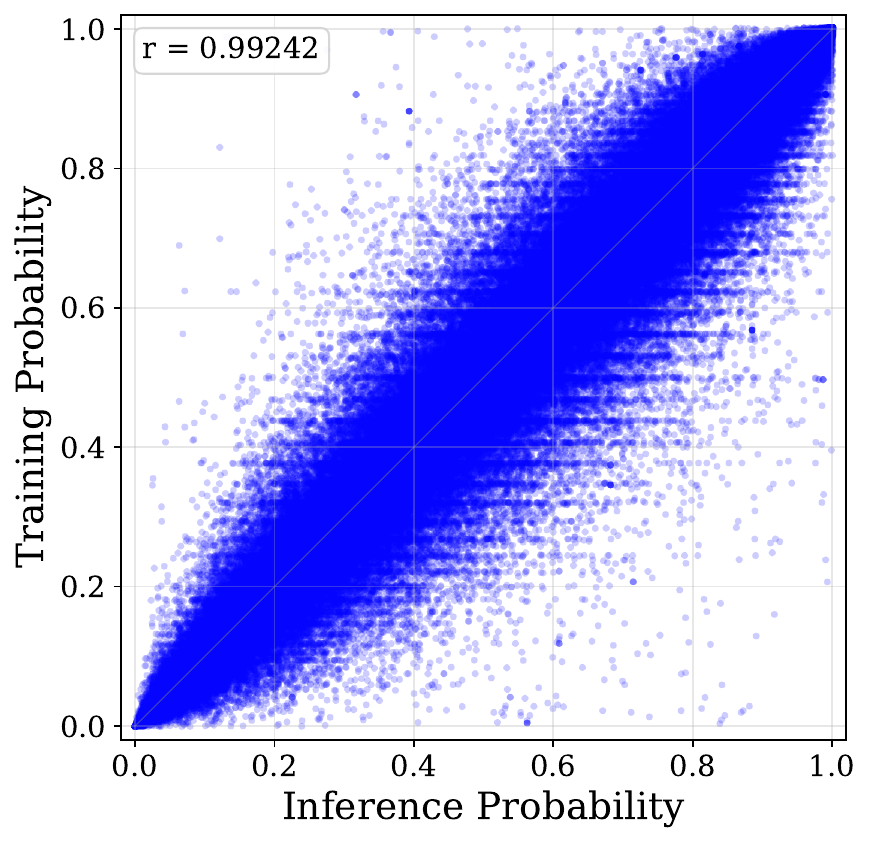}
        \centerline{\small (c) Log-Prob Diff}
    \end{minipage}
    \caption{Naive NVFP4 collapse in RL.
    \textbf{(a)}~Mean training reward: naive NVFP4 W4A4 rollout with BF16 training (blue) peaks near step 150 and then collapses, whereas the BF16 RL baseline (red) continues to improve steadily.
    \textbf{(b)}~Held-out test score shows the same pattern, confirming that the failure is not limited to the training reward signal.
    \textbf{(c)}~Log-probability difference between the NVFP4 rollout engine and the BF16 trainer diverges extremely, indicating a severe training--inference mismatch that importance sampling cannot correct.}
    \label{fig:intro-failure}
\end{figure}

Recently, NVIDIA Blackwell introduces a more aggressive target.
NVFP4 enables native \emph{W4A4} general matrix multiplications (GEMMs) on FP4 Tensor Cores, with peak GEMM throughput up to $\sim$2$\times$ higher than FP8 and $\sim$4$\times$ over BF16~\citep{alvarez2025nvfp4blog,jarmusch2025blackwell,nvidia2025nvfp4pretrain}.
However, NVFP4 is qualitatively less forgiving than FP8: its E2M1 core has only eight positive representable levels per sign, compared with 256 for FP8 E4M3, and true W4A4 execution quantizes activations on every rollout forward pass.
As shown in Figure~\ref{fig:intro-failure}, directly combining NVFP4 W4A4 rollout with BF16 training collapses within roughly 150 steps: reward and held-out score both fall while the per-token log-probability gap grows monotonically, pushing importance ratios outside the trust region.

To identify the source of this failure, we conduct a motivating ablation (Section~\ref{sec:motivating-exp}) that independently quantizes weights and activations during rollout.
Comparing W4A16 (FP4 weights, BF16 activations) with W16A4 (BF16 weights, FP4 activations) reveals a strong operand asymmetry: \emph{activation quantization}, not weight quantization, is the dominant driver of log-probability divergence and reward collapse.
W16A4 closely tracks the failure mode of full W4A4, whereas W4A16 remains much closer to stable behavior; the two error sources are approximately additive.
This reframes NVFP4 RL as a quantization error alignment problem.

Existing approaches do not resolve this setting.
Weight-only NVFP4 methods such as QeRL~\citep{huang2025qerl} keep activations in BF16 and execute FP16 GEMMs instead of native W4A4 FP4 Tensor Core operations, while FP8 and INT8 based RL frameworks~\citep{li2026qurl,gu2026qarl,xi2026jetrl,qiu2026fp8rl,zhou2026ais} operate on finer numerical grids where rollout--trainer mismatch is more tractable.
Stable full-parameter MoE GRPO with true W4A4 rollout on Blackwell therefore still requires alignment strategies that directly target the activation-dominated mismatch identified above.

We address this gap with QUantization-error Alignment across Dual Sides (QUADS) for NVFP4 RL.
On the training side, asymmetric W4A16 quantization-aware training aligns the weight quantization--dequantization path while keeping learner activations in BF16 (Section~\ref{sec:methods:w4a16-qat}).
This design exploits the fact that weights are synchronized across engines and can therefore share a matched QDQ path, while activations are recomputed online and cannot be reliably aligned by symmetric W4A4 fake quantization.
On the rollout side, residual activation compensation reduces the remaining rollout-time activation error through a second-pass quantization of high-residual channels, implemented with fused kernels to preserve most of the W4A4 throughput advantage (Section~\ref{sec:methods:residual}).
Together, these components recover BF16-level accuracy in full-parameter MoE GRPO: our QUADS pipeline reaches 72.86\% average pass@1 across four held-out benchmarks, matching the 73.15\% BF16 baseline and avoiding the severe degradation of naive NVFP4 training (51.37\%). At the same time, it improves rollout throughput by 16\% over FP8. 

Our main contributions are:
\begin{itemize}[leftmargin=*,itemsep=2pt,topsep=2pt]
    \item \textbf{NVFP4 RL failure diagnosis.}
    Through ablations, we show that activation quantization is essential in NVFP4 RL stability and explains the collapse of naive W4A4 RL.
    \item \textbf{Asymmetric Quantization-Aware-Training.}
    On the trainer-side, we propose the asymmetric QAT scheme that fake-quantizes weights to NVFP4 but leaves activations unquantized, reducing weight mismatch without amplifying cross-engine activation drift.
    \item \textbf{Residual Activation Compensation.}
    On the rollout-side, we further reduce activation error through second-pass quantization of high-residual channels, preserving native W4A4 GEMMs while closing the remaining log-probability gap.
    \item \textbf{NVFP4 RL empirical validation for MoE.}
    On NVFP4 RL training of MoE model, our proposed QUADS pipeline matches BF16-level benchmark accuracy and retains a throughput advantage over FP8 rollout.
\end{itemize}

\section{Preliminaries}
\label{sec:preliminaries}

\subsection{Reinforcement Learning for LLMs}
\label{sec:rl-llm}

Modern LLM post-training increasingly relies on Reinforcement Learning with Verifiable Rewards (RLVR).
A typical pipeline decouples \emph{rollout}---autoregressive response generation on a throughput-oriented inference engine, from \emph{training}---gradient computation on a separate backend such as FSDP or Megatron.
Policy-gradient algorithms update an autoregressive policy $\pi_\theta$ using samples drawn from a reference policy $\pi_{\mathrm{ref}}$.
Throughout this work we adopt GRPO~\citep{shao2024deepseekmath} as the representative algorithm and analyze how numerical precision in rollout affects its stability.

\paragraph{GRPO}

Group Relative Policy Optimization (GRPO)~\citep{shao2024deepseekmath} extends PPO~\citep{schulman2017ppo} by eliminating the value model: for each prompt, it samples a group of $G$ responses $\{o_i\}_{i=1}^{G}$ from the old policy $\pi_{\theta_{\mathrm{old}}}$ and normalizes their scalar rewards within the group to obtain advantages $\hat{A}_{i,t}$.
The clipped surrogate objective is
\begin{equation}
    \mathcal{J}_{\mathrm{GRPO}}(\theta) =
    \mathbb{E}\!\left[
        \frac{1}{G}\sum_{i=1}^{G}\frac{1}{|o_i|}\sum_{t=1}^{|o_i|}
        \min\!\Bigl(
            r_{i,t}(\theta)\,\hat{A}_{i,t},\;
            \mathrm{clip}\bigl(r_{i,t}(\theta),\, 1-\epsilon,\, 1+\epsilon\bigr)\,\hat{A}_{i,t}
        \Bigr)
        - \beta\, D_{\mathrm{KL}}(\pi_{\theta}\,\|\,\pi_{\mathrm{ref}})
    \right],
    \label{eq:grpo}
\end{equation}
where $r_{i,t}(\theta)$ is the per-token importance ratio (Eq.~\eqref{eq:is-ratio}), $\epsilon$ bounds the trust region, and $\beta$ controls the KL penalty against $\pi_{\mathrm{ref}}$.
Both the clipped surrogate and the KL term depend on token-level log-probabilities; biased estimates therefore directly corrupt the policy gradient.

\paragraph{Training--Inference Mismatch}
\label{sec:rl-mismatch}

When rollout and training share the same engine and numerical path, GRPO reduces to standard on-policy optimization.
In practice, however, responses are sampled by an inference policy $\pi_{\mathrm{sampler}}$ (e.g., vLLM or SGLang) while gradients are computed with a training policy $\pi_{\mathrm{learner}}$ (e.g., FSDP)~\citep{yao2025offpolicy}.
Importance sampling (IS) corrects this off-policy gap: for each token $a_t$ given context $s_t$, the per-token ratio is
\begin{equation}
    r_t(\theta) = \frac{\pi_{\theta}(a_t \mid s_t)}{\pi_{\theta_{\mathrm{old}}}(a_t \mid s_t)}
    = \exp\!\bigl(\log \pi_{\theta}(a_t \mid s_t) - \log \pi_{\theta_{\mathrm{old}}}(a_t \mid s_t)\bigr).
    \label{eq:is-ratio}
\end{equation}
We denote the log-probability difference $\delta_t \triangleq \log \pi_{\theta}(a_t \mid s_t) - \log \pi_{\theta_{\mathrm{old}}}(a_t \mid s_t)$.
When $\pi_{\theta_{\mathrm{old}}}$ and $\pi_{\theta}$ are evaluated at different precisions or through different kernel paths, $\delta_t \neq 0$ even for identical parameters---this is the \emph{training--inference mismatch}~\citep{qi2025precisionrl,gu2026qarl,qiu2026fp8rl}.

Low-precision rollout amplifies the problem.
In FP8 settings the resulting gaps are comparatively modest and can often be controlled with truncated importance sampling (TIS)~\citep{yao2025offpolicy}, adaptive clipping~\citep{li2026qurl,zhou2026ais}, or unified FP8 precision flows~\citep{xi2026jetrl,qiu2026fp8rl}.
Under NVFP4 W4A4 rollout with BF16 training, by contrast, $|\delta_t|$ grows rapidly over optimization steps, importance ratios leave $[1-\epsilon,\, 1+\epsilon]$, and reward curves collapse (Figure~\ref{fig:intro-failure}).

We analyze mismatch through the decomposition
\begin{equation}
    \delta_t \;\approx\; \delta_t^{\mathrm{weight}} + \delta_t^{\mathrm{act}} + \delta_t^{\mathrm{num}},
    \label{eq:decomposition}
\end{equation}
where $\delta_t^{\mathrm{weight}}$ arises from weight quantization, $\delta_t^{\mathrm{act}}$ from activation quantization, and $\delta_t^{\mathrm{num}}$ from residual numerical-path differences between engines~\citep{qi2025precisionrl,yao2025offpolicy}.
This decomposition guides our dual-side alignment strategy in Section~\ref{sec:methods}.

\subsection{Low-Precision Inference}
\label{sec:low-precision}

Because rollout dominates RL wall-clock time, a standard acceleration strategy is to execute linear-layer GEMMs at reduced precision while keeping gradient computation in BF16 or FP32.
Quantization maps a high-precision tensor $X$ to a low-precision surrogate $\hat{X}$ via a scale factor and a discrete rounding operator.
Following the unified formulation in~\citep{li2026qurl}, a $b$-bit quantized value can be written as
\begin{equation}
    Q(X;\, b,\, e,\, \alpha)
    = \alpha \cdot (-1)^{\mathrm{sign}}
    \cdot 2^{d}
    \cdot \Bigl(1 + \sum_{i=1}^{b-1-e} \frac{m_i}{2^{i}}\Bigr),
    \label{eq:quant-unified}
\end{equation}
where $\mathrm{sign} \in \{0,1\}$, the exponent $d$ uses $e$~bits, each mantissa bit $m_i \in \{0,1\}$, and the scaling factor $\alpha$ is determined by the dynamic range of a quantization group (per-tensor, per-channel, or per-block).
In practice, inference stacks apply Eq.~\eqref{eq:quant-unified} independently to weights and activations before GEMM; we denote a $b$-bit weight--activation configuration as W$b$A$b'$ when both operands use $b$ and $b'$ bits, respectively.

Floating-point quantization keeps a non-zero exponent field ($e > 0$ in Eq.~\eqref{eq:quant-unified}), yielding a non-uniform grid that concentrates representable levels near zero and preserves wider dynamic range at the cost of irregular spacing between levels.

\paragraph{FP8.}
The E4M3 format used in LLM inference sets $b=8$, $e=4$, and three mantissa bits, providing 256 positive representable levels per sign and a maximum representable magnitude of $\Delta_{\max}=448$.
FP8 W8A8 GEMMs are natively supported on Hopper and Blackwell FP8 Tensor Cores and underpin FP8 rollout in recent RL systems~\citep{xi2026jetrl,qiu2026fp8rl,li2026qurl}.
DeepGEMM~\citep{deepgemm2025} implements high-throughput FP8 GEMMs with \emph{fine-grained} block scaling: rather than a single scale per tensor, each tile of values carries its own $\alpha$ in Eq.~\eqref{eq:quant-unified}, matching the block-wise layouts of large MoE models more closely than per-tensor schemes.
The library JIT-compiles kernels at runtime (\texttt{fp8\_gemm\_nt}, \texttt{m\_grouped\_fp8\_gemm\_*}) and is widely used for FP8 inference prefilling, making FP8 rollout a mature acceleration path for RL.

\paragraph{FP4.}
\label{sec:nvfp4}
Where FP8 E4M3 already trades a 256-level grid for roughly $2\times$ GEMM throughput over BF16, FP4 pushes the same trade-off one step further: halving bit width again targets roughly $4\times$ throughput, but shrinks the representable set to only eight positive magnitudes per sign under a typical E2M1 core ($b=4$, $e=2$, one mantissa bit)---$32\times$ coarser than E4M3.
The resulting quantization error is larger and less uniformly distributed, so FP4 rollout amplifies the training--inference mismatch of Section~\ref{sec:rl-mismatch} beyond what FP8 IS corrections typically absorb.
NVFP4 is NVIDIA's block-scaled instantiation of this precision tier, native to Blackwell FP4 Tensor Cores~\citep{alvarez2025nvfp4blog,nvidia2025nvfp4pretrain,jarmusch2025blackwell}.
In Eq.~\eqref{eq:quant-unified}, the NVFP4 \emph{core} uses E2M1 encoding and yields $\{0, 0.5, 1, 1.5, 2, 3, 4, 6\}$ per sign; sixteen consecutive elements share an E4M3 \emph{block scale} that sets their common $\alpha$, and an optional per-tensor FP32 scale provides a second level of dynamic-range control~\citep{alvarez2025nvfp4blog}.

\section{Methods}
\label{sec:methods}

Naive NVFP4 W4A4 rollout with BF16 training fails because low-precision inference induces a large training--inference mismatch that destabilizes optimization.
This section develops a dual-side alignment strategy.
Section~\ref{sec:motivating-exp} first models the training--inference forward error and shows experimentally that activation FP4 quantization is the dominant and harder-to-align source of the log-probability gap between rollout and learner.
Section~\ref{sec:methods:w4a16-qat} then derives asymmetric W4A16 QAT: for trainer-side, we fake-quantize weights to FP4, but deliberately keep training activations in BF16.
Section~\ref{sec:methods:residual} finally reduces the remaining rollout-side activation error through targeted residual compensation.

\begin{table}[htbp]
\centering
\caption{Training-side QDQ has opposite effects on weight and activation mismatch.
Weight QDQ aligns W4A16 rollout with the learner, whereas activation QDQ does not shrink W16A4 mismatch because the two engines quantize already-different activations.}
\label{tab:qdq-alignment}
\begin{tabular}{lccp{0.42\linewidth}}
\toprule
Rollout & Learner & Targeted term & Observed effect \\
\midrule
W4A16 & BF16 & $\Delta\mathbf{W}_q$ & Large training--inference log-probability gap \\
W4A16 & W4A16 QAT & $\Delta\mathbf{W}_q$ & Gap is significantly reduced \\
W16A4 & BF16 & $\boldsymbol{\eta}+\Delta\mathbf{X}_q$ & Large training--inference log-probability gap \\
W16A4 & W16A4 QAT & $\boldsymbol{\eta}+\Delta\mathbf{X}_q$ & Gap remains large \\
\bottomrule
\end{tabular}
\end{table}

\subsection{FP4 RL Mismatch Error Analysis}
\label{sec:motivating-exp}

Figure~\ref{fig:intro-failure} shows the symptom: W4A4 rollout with a BF16 learner collapses within roughly 150 optimization steps, while BF16 rollout remains stable.
For a fixed checkpoint and token sequence, let $\pi_{\theta_{\mathrm{old}}}$ be the rollout policy evaluated by the inference engine and $\pi_{\theta}$ be the learner policy evaluated by the training engine.
We write the per-token gap $\delta_t$ at a coarse level as
\begin{equation}
    \delta_t
    \;\approx\;
    \delta_t^{\mathrm{weight}}
    + \delta_t^{\mathrm{act}}
    + \delta_t^{\mathrm{num}},
    \label{eq:delta-decomp-methods}
\end{equation}
where the three terms correspond to weight quantization, activation quantization, and numerical differences between decoupled engines~\citep{qi2025precisionrl,yao2025offpolicy}.
The key question is which term can be reduced by standard quantization--dequantization (QDQ) alignment, and which term remains exposed in true W4A4 rollout.

\begin{figure}[t]
    \centering
    \begin{minipage}{0.48\linewidth}
        \centering
        \includegraphics[width=\linewidth]{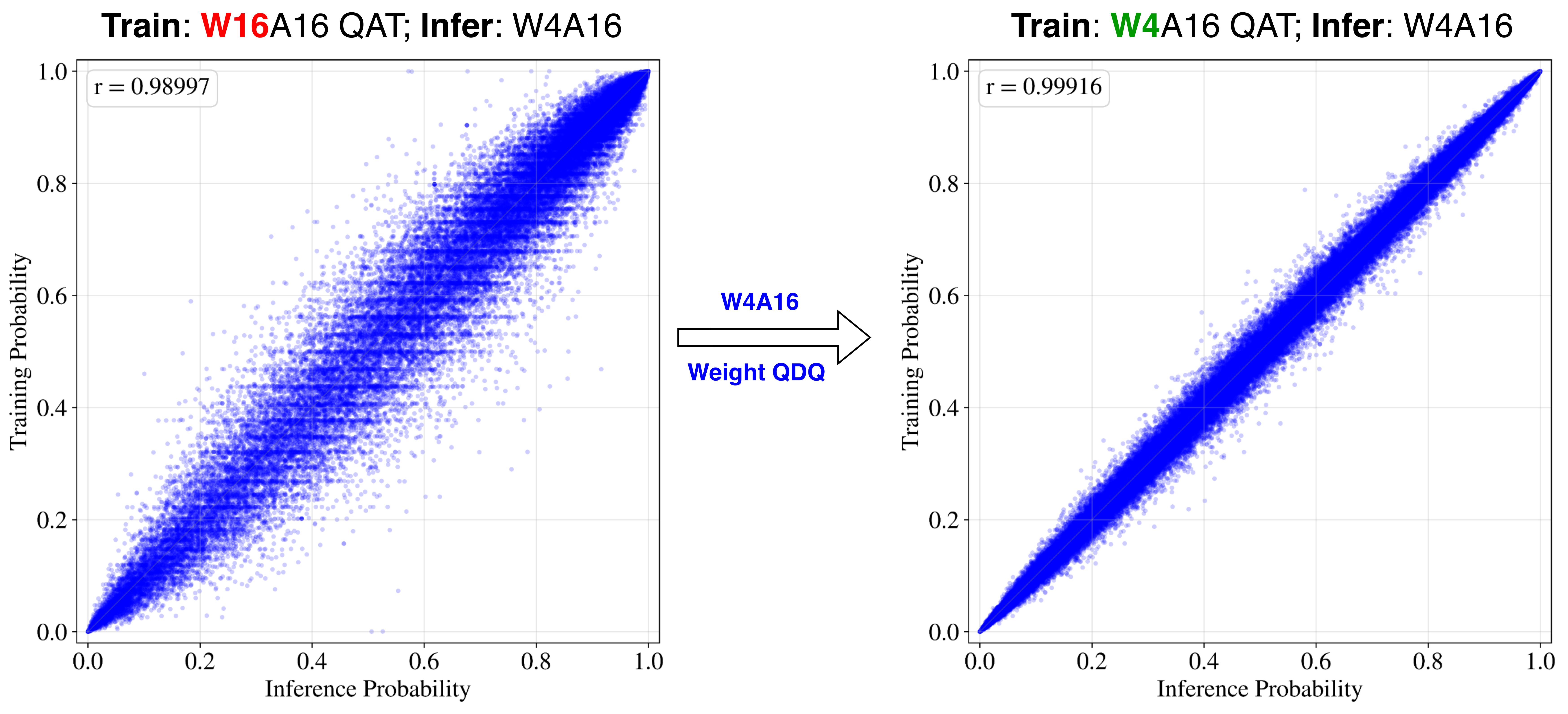}
        \centerline{\small (a) Weight QDQ Alignment}
    \end{minipage}\hfill
    \begin{minipage}{0.48\linewidth}
        \centering
        \includegraphics[width=\linewidth]{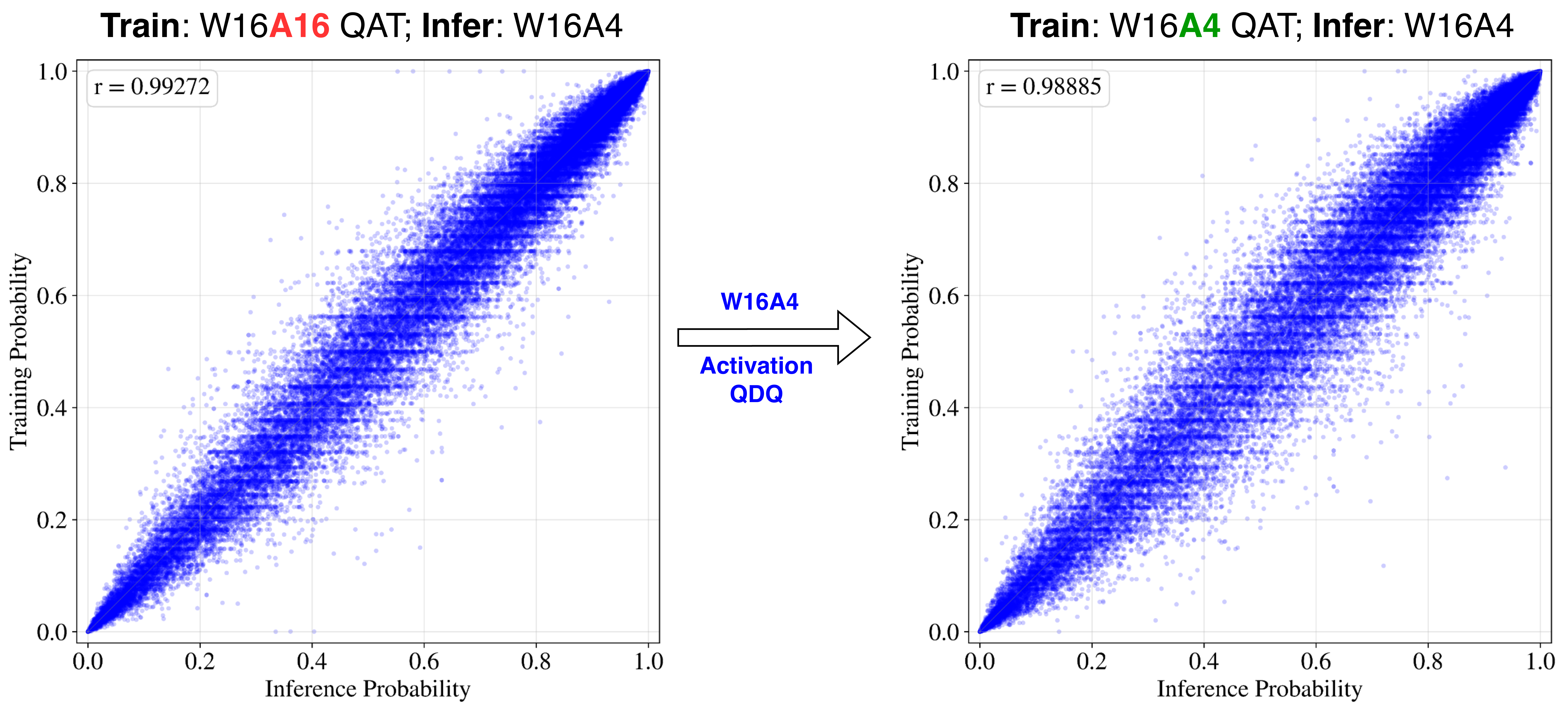}
        \centerline{\small (b) Activation QDQ Alignment}
    \end{minipage}
    \caption{Per-token log-probability gap under the motivating ablation.
    W16A4 is much closer to W4A4 than to W4A16, identifying activation quantization as the dominant source of mismatch.
    Weight QDQ reduces the W4A16 gap while activation QDQ does not reduce the W16A4 gap.}
    \label{fig:motivating-diff}
\end{figure}

\paragraph{Layer-wise source of the mismatch.}
Consider one linear or MoE projection and let the BF16 training output be
\begin{equation}
    \mathbf{Y}_{\mathrm{train}}
    =
    \mathbf{W}_{\mathrm{train}}\mathbf{X}_{\mathrm{train}} .
    \label{eq:train-linear}
\end{equation}
The inference engine receives synchronized weights but recomputes activations through a different numerical path before applying NVFP4 quantization.
We denote the weight quantization error by
\begin{equation}
    \Delta\mathbf{W}_q
    =
    Q_W(\mathbf{W}_{\mathrm{infer}})
    -
    \mathbf{W}_{\mathrm{infer}},
    \label{eq:weight-error-def}
\end{equation}
where $Q_W$ is the hardware-aligned NVFP4 weight QDQ operator.
For activations, the inference-side BF16 activation differs from the learner activation even before FP4 quantization:
\begin{equation}
    \boldsymbol{\eta}
    =
    \mathbf{X}_{\mathrm{infer}}
    -
    \mathbf{X}_{\mathrm{train}} .
    \label{eq:engine-eta-def}
\end{equation}
After activation quantization,
\begin{equation}
    \Delta\mathbf{X}_q
    =
    Q_A(\mathbf{X}_{\mathrm{infer}})
    -
    \mathbf{X}_{\mathrm{infer}},
    \label{eq:act-error-def}
\end{equation}
where $Q_A$ is the NVFP4 activation QDQ operator.
The inference output can therefore be written as
\begin{equation}
    \mathbf{Y}_{\mathrm{infer}}
    =
    (\mathbf{W}_{\mathrm{train}}+\Delta\mathbf{W}_q)
    (\mathbf{X}_{\mathrm{train}}+\boldsymbol{\eta}+\Delta\mathbf{X}_q).
    \label{eq:infer-linear}
\end{equation}
Subtracting Eq.~\eqref{eq:train-linear} and taking the Frobenius norm defines the layer-wise forward mismatch
\begin{equation}
    \Delta\mathbf{Y}
    \triangleq
    \left\|\mathbf{Y}_{\mathrm{infer}}-\mathbf{Y}_{\mathrm{train}}\right\|_2
    =
    \left\|
    \mathbf{W}_{\mathrm{train}} \star \underbrace{(\boldsymbol{\eta}+\Delta\mathbf{X}_q)}_{\mathrm{activation\ error}}
    +
    \underbrace{\Delta\mathbf{W}_q}_{\mathrm{weight\ error}} \star \mathbf{X}_{\mathrm{train}}
    +
    \underbrace{\Delta\mathbf{W}_q(\boldsymbol{\eta}+\Delta\mathbf{X}_q)}_{\mathrm{second\ order}}
    \right\|_2 .
    \label{eq:linear-mismatch}
\end{equation}
Equation~\eqref{eq:linear-mismatch} is the central distinction.
The weight error is solely a quantization error, while the activation error contains not only the FP4 quantization error $\Delta\mathbf{X}_q$, but also the engine mismatch $\boldsymbol{\eta}$; 
Since the inference side weights are replicated from the training side, ensuring $\mathbf{W}_{\mathrm{infer}}= \mathbf{W}_{\mathrm{train}}$, applying the same quantization function to the weights on the training side is sufficient to eliminate the weight error, i.e., $\|\Delta\mathbf{W}_q\|_2 = 0$.
However, activation error is hard to align. In Figure~\ref{fig:motivating-diff}, we conduct two experiments, namely Weight QDQ Alignment and Activation QDQ Alignment, which illustrate QDQ alignment only works for weights, but not for activations.

\subsection{Asymmetric Quantization-Aware Training for Training-side}
\label{sec:methods:w4a16-qat}

Based on the above experimental observations, in the setting of FP4 W4A4 RL, although rollout uses W4A4, we propose an asymmetric QAT on the training-side, that is, perform FP4 QAT on the weights on the trainer, and maintain BF16 for the activations.

\paragraph{Forward and backward of Asymmetric QAT.}
For each linear or MoE projection, let $\mathrm{QDQ}_W(\cdot)$ denote the FP4 weight fake-quantization operator whose scale layout and rounding match the rollout kernel.
The asymmetric QAT forward pass is
\begin{equation}
    \widehat{\mathbf{W}}
    =
    \mathrm{QDQ}_W(\mathbf{W}),
    \qquad
    \mathbf{Y}_{\mathrm{qat}}
    =
    \widehat{\mathbf{W}}\mathbf{X}_{\mathrm{train}}^{\mathrm{BF16}},
    \label{eq:w4a16-qat-forward}
\end{equation}
with no activation fake quantization on the learner.
The backward pass uses the straight-through estimator (STE) for the QDQ node:
\begin{equation}
    \frac{\partial\mathcal{L}}{\partial \mathbf{W}}
    \approx
    \frac{\partial\mathcal{L}}{\partial \widehat{\mathbf{W}}},
    \qquad
    \frac{\partial\mathcal{L}}{\partial \mathbf{X}_{\mathrm{train}}}
    =
    \widehat{\mathbf{W}}^{\top}
    \frac{\partial\mathcal{L}}{\partial \mathbf{Y}_{\mathrm{qat}}}.
    \label{eq:w4a16-qat-backward}
\end{equation}
Thus the optimizer still updates BF16 master weights, but every learner forward pass sees the same discretized weight values used by rollout.

To demonstrate why trainer-side W4A16 QAT is superior, we conduct the following analysis.
To make the contrast explicit, let $\Delta\mathbf{Y}^{\mathrm{pre}}$ and $\Delta\mathbf{Y}^{\mathrm{post}}$ denote the layer-wise mismatch norm before and after inserting the corresponding QDQ node into the learner forward pass.

\paragraph{Why weights should be quantized on the learner.} We use the weight-only quantization setting to demonstrate the effect of weight error.  
In this setting, activations remain BF16 and only the weight term in Eq.~\eqref{eq:linear-mismatch} is active, then the mismatch is
\begin{equation}
    \Delta\mathbf{Y}^{\mathrm{pre}}
    =
    \left\|
    \mathbf{W}_{\mathrm{train}}\star \boldsymbol{\eta}
    +
    \Delta\mathbf{W}_q \star \mathbf{X}_{\mathrm{train}}
    +
    \Delta\mathbf{W}_q\boldsymbol{\eta}
    \right\|_2.
    \label{eq:w4a16-pre-qdq}
\end{equation}
After utilizing weight fake-quantization on the learner, both sides multiply by the same quantized weight. This is because the rollout engines load a bit-exact copy of the trainer parameters at each step:
\begin{equation}
    \mathbf{W}_{\mathrm{infer}} = \mathbf{W}_{\mathrm{train}}.
    \label{eq:weight-sync}
\end{equation}
If the training-side quantization $Q_W^{'}$ matches the rollout-side, then the weight quantization error vanishes:
\begin{equation}
    \left\|
        \Delta\mathbf{W}_q
    \right\|_2
    =
    \left\|
        Q_W(\mathbf{W}_{\mathrm{infer}})
        -
        Q_W^{'}(\mathbf{W}_{\mathrm{train}})
    \right\|_2
    =
    0.
    \label{eq:weight-qdq-align}
\end{equation}

So the train--inference mismatch becomes $\Delta\mathbf{Y}^{\mathrm{post}}=\left\|\mathbf{W}_{\mathrm{train}} \star \boldsymbol{\eta}\right\|_2$.
Since $\Delta\mathbf{W}_q\mathbf{X}_{\mathrm{train}}$ is generally nonzero, then weight QDQ is guaranteed to shrink the mismatch:
\begin{equation}
    \Delta\mathbf{Y}^{\mathrm{post}}
    <
    \Delta\mathbf{Y}^{\mathrm{pre}}.
    \label{eq:w4a16-qdq-improves}
\end{equation}

\begin{figure}[htbp]
    \centering
    \begin{minipage}{0.48\linewidth}
        \centering
        \includegraphics[width=\linewidth]{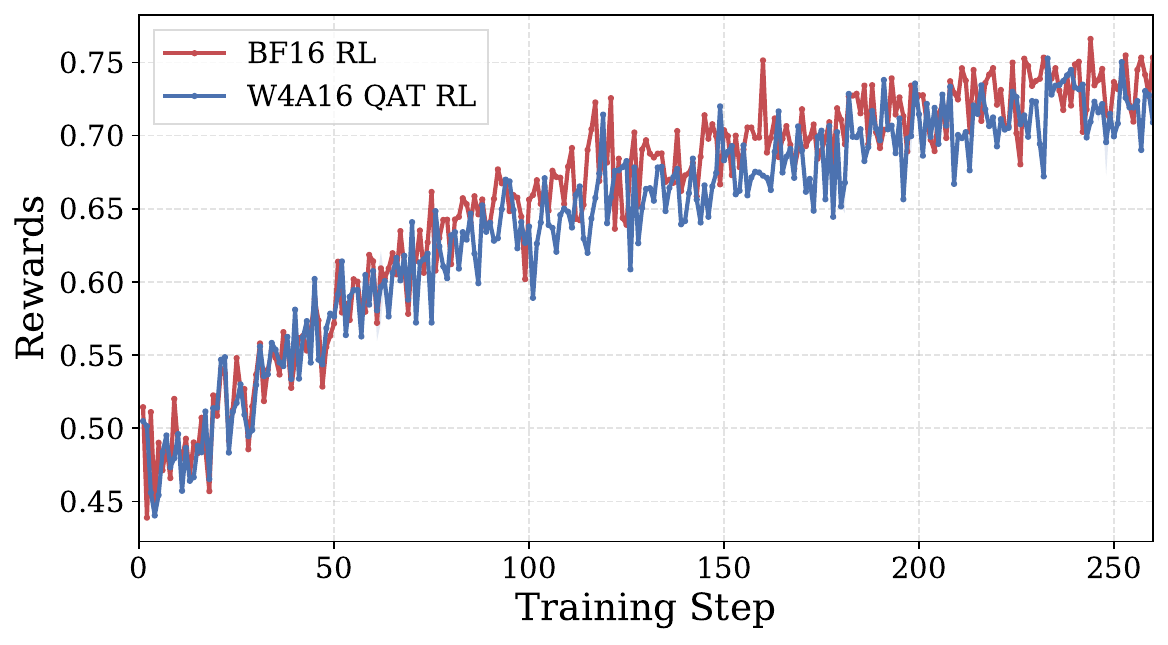}
        \centerline{\small (a) Training Reward}
    \end{minipage}\hfill
    \begin{minipage}{0.48\linewidth}
        \centering
        \includegraphics[width=\linewidth]{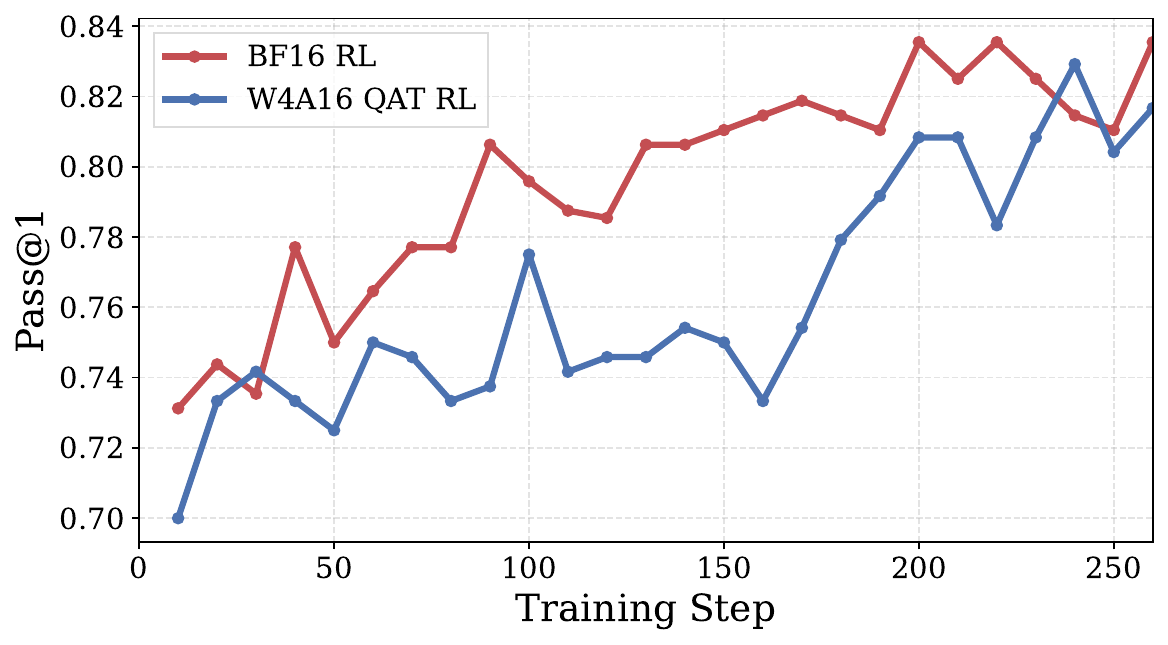}
        \centerline{\small (b) AIME24 Test Score}
    \end{minipage}
    \caption{W4A16 QAT improves over naive W4A4 but does not match BF16 RL stability.
    \textbf{(a)}~Mean training reward: W4A16 QAT with W4A4 inference (red) avoids collapse yet grows more slowly than the BF16 rollout baseline (blue).
    \textbf{(b)}~AIME24 test score shows the same gap: red remains below blue during almost all the training.}
    \label{fig:qat-only-failure}
\end{figure}

\textbf{Why activation should remain unquantized on the learner.}
We use the activation-only quantization setting to demonstrate the effect of activation error.
In this setting, weights remain BF16 and only the activation term in Eq.~\eqref{eq:linear-mismatch} is active, then the mismatch is
\begin{equation}
    \Delta\mathbf{Y}^{\mathrm{pre}}
    =
    \left\|
    \mathbf{W}_{\mathrm{train}}(\boldsymbol{\eta}+\Delta\mathbf{X}_q)
    \right\|_2.
    \label{eq:w16a4-pre-qdq}
\end{equation}
After inserting activation QDQ on the learner, the comparison becomes
\begin{equation}
    \Delta\mathbf{Y}^{\mathrm{post}}
    =
    \left\|
    \mathbf{W}_{\mathrm{train}}
    \bigl[
        Q_A(\mathbf{X}_{\mathrm{train}}+\boldsymbol{\eta})
        -
        Q_A(\mathbf{X}_{\mathrm{train}})
    \bigr]
    \right\|_2,
    \label{eq:act-qdq-mismatch}
\end{equation}
because the two engines now apply $Q_A$ to already-different activations.
Unlike the weight case, applying activation QDQ does not guarantee to reduce and can even widen the training--inference gap.
This situation is verified in Figure~\ref{fig:w4a16-vs-w4a4-qat}, the activation QDQ does not reduce the training--inference gap, and even increases it.

\subsection{Activation Error Compensation for Rollout-side}
\label{sec:methods:residual}
Although W4A16 QAT on the training-side can reduce some train-inference mismatch to avoid the catastrophic collapse of naive FP4 RL, it is still not enough to match the BF16 baseline.
As shown in Figure~\ref{fig:qat-only-failure}, W4A16 QAT with W4A4 rollout avoids the catastrophic collapse, but its reward grows more slowly and its evaluation score remains below the BF16 rollout baseline.

The remaining bottleneck is the activation part of Eq.~\eqref{eq:linear-mismatch}.
Only considering the training side cannot reduce the activation error within it. To minimize the activation error, we continue from the inference side and consider the activation quantization error $\Delta \mathbf{X}_q$ of it.

\begin{figure}[htbp]
    \centering
    \begin{minipage}{0.45\linewidth}
        \centering
        \includegraphics[width=\linewidth]{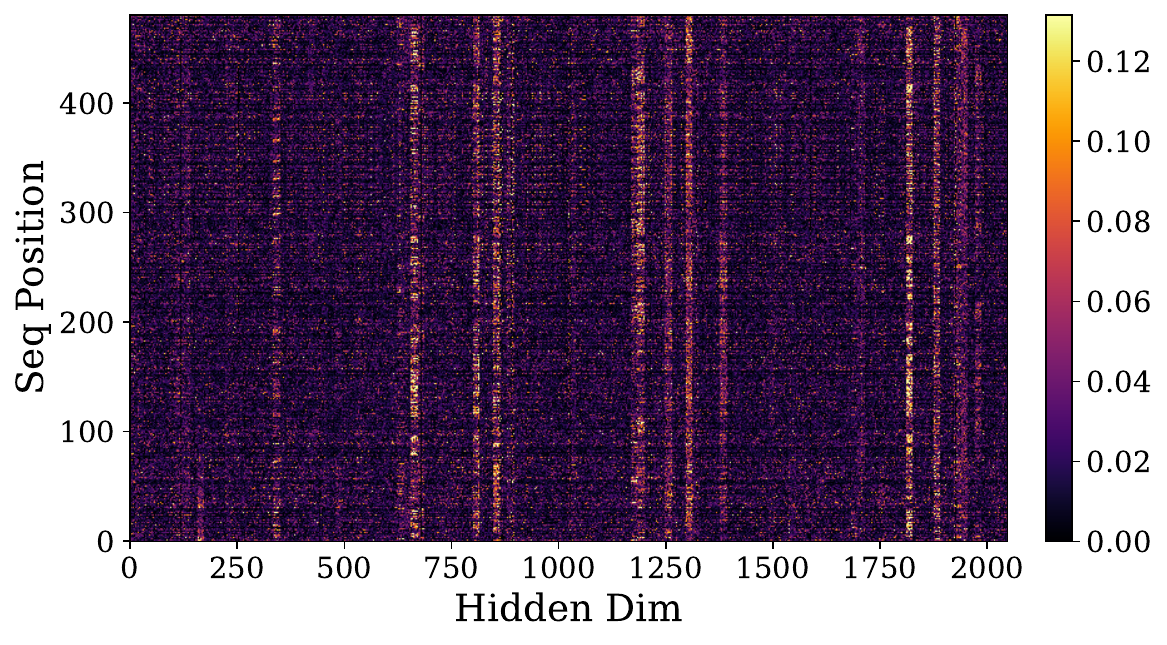}
        \caption*{\textbf{(a)} NVFP4 activation quantization residual motivates top-k\% channel selection.}
    \end{minipage}\hfill
    \begin{minipage}{0.45\linewidth}
        \centering
        \includegraphics[width=\linewidth]{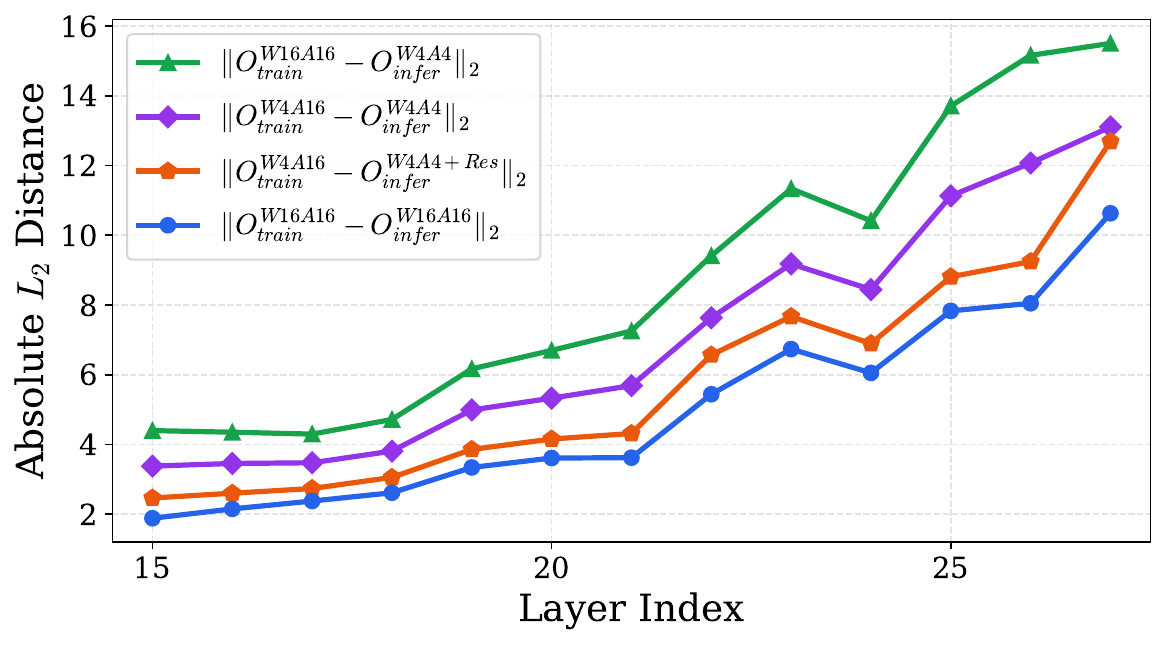}
        \caption*{\textbf{(b)} Layer-wise MoE-output mismatch under four training--inference configurations.}
        \label{fig:w4a16-w16a4-output}
    \end{minipage}
    \caption{\textbf{(a)} Absolute NVFP4 activation quantization residual  $|\Delta \mathbf{X}_q| = |Q_A(\mathbf{X}) - \mathbf{X}|$  on a representative MoE activation (color scale clipped at the 99.5th percentile for visibility).
    Residual magnitude varies sharply across hidden dimensions: most channels remain near zero, while a small set of outlier channels form persistent vertical bands across sequence positions, motivating top-$k$\% channel selection for targeted compensation.
    \textbf{(b)} Layer-wise MoE output $L_2$ distance under four training--inference configurations. Without QAT (green, W16A16 train / W4A4 infer), the output mismatch is largest. W4A16 QAT (purple) reduces the gap by aligning weights to the quantized grid. Adding residual compensation (orange, W4A4+Res infer) further closes the distance toward the BF16 baseline (blue), confirming that both weight-side QAT and activation-side compensation contribute to reducing the training--inference mismatch progressively across all layers.}
    \label{fig:residual-distribution}
    \label{fig:dual-side-alignment}
\end{figure}

For a BF16 activation $\mathbf{X}$, the definition of the activation quantization error is as follows:
\begin{equation}
    \Delta \mathbf{X}_q
    =
    Q_A(\mathbf{X})
    -
    \mathbf{X},
    \label{eq:residual-def}
\end{equation}
which measures the error introduced by FP4 activation quantization.
This quantization error is not uniform across hidden dimensions.
Because block scales are determined by local dynamic range, some high-magnitude channels create large quantization errors, while other channels have much smaller quantization errors.
Figure~\ref{fig:residual-distribution} shows this per-channel structure: high-residual channels appear as persistent vertical bands.
This motivates allocating additional computation only to channels with large $\|\Delta \mathbf{X}_q\|$.

\paragraph{Targeted residual compensation.}
Rather than using $Q_A(\mathbf{X})$ directly, we correct the quantization error on a selected channel set $\mathcal{S}$:
\begin{equation}
    \tilde{\mathbf{X}}
    \;=\;
    Q_A(\mathbf{X})
    -
    \mathbf{M}_{\mathcal{S}}
    \odot
    Q_A(\Delta \mathbf{X}_q),
    \label{eq:residual-recon}
\end{equation}
where $\mathbf{M}_{\mathcal{S}}$ is a binary channel mask and $\odot$ denotes element-wise multiplication.
Channels outside $\mathcal{S}$ follow the standard W4A4 path.
Channels inside $\mathcal{S}$ receive a second FP4 correction term, reducing the distance to the BF16 reference:
\begin{equation}
    \bigl\|\tilde{\mathbf{X}} - \mathbf{X}_{\mathrm{train}}\bigr\|_2
    \;<\;
    \bigl\|Q_A(\mathbf{X}) - \mathbf{X}_{\mathrm{train}}\bigr\|_2,
    \label{eq:residual-shrink}
\end{equation}
thereby shrinking the activation contribution to the log-probability gap from the inference side.
The correction is residual-based rather than full BF16 fallback, so the main GEMM remains on the W4A4 FP4 Tensor Core path.

\paragraph{Online per-channel selection.}
RL changes the policy every update, so activation statistics are non-stationary.
We therefore choose $\mathcal{S}$ online for each forward pass.
For channel $j$, define the residual score
\begin{equation}
    s_j
    =
    \left\|
        \Delta \mathbf{X}_q[:,j]
    \right\|_2,
    \label{eq:channel-score}
\end{equation}
aggregated over the current token batch, and keep the largest $k\%$ channels:
\begin{equation}
    \mathcal{S}
    =
    \operatorname{TopK}_{k\%}
    \left(
        \{s_j\}_{j=1}^{d}
    \right).
    \label{eq:channel-select}
\end{equation}
In our implementation we set $k=50$ for the profiled W13 activation site, which captures the large residual channels while adding only about $\sim$10\% additional FLOPs relative to plain W4A4 rollout.
With W4A16 QAT plus this rollout-side compensation, the resulting training curve closely matches BF16 rollout and training.
Figure~\ref{fig:dual-side-alignment}(b) confirms this at the tensor level: the per-layer MoE output mismatch under W4A16 QAT with residual compensation (orange) approaches the BF16 baseline (blue), whereas W4A16 QAT alone (purple) or no QAT (green) leaves a substantially larger gap.

\paragraph{Efficient implementation.}
A naive implementation of Eqs.~\eqref{eq:residual-def}--\eqref{eq:residual-recon} launches separate kernels for activation quantization, residual extraction, channel selection, residual quantization, masking, and reconstruction, which would erase much of the W4A4 speed benefit.
We fuse the dominant stages into a Triton kernel, reducing launch overhead and keeping the end-to-end cost near the $\sim$10\% FLOP increase from the selected residual channels.
Together, asymmetric W4A16 QAT and inference-side residual compensation form the intended dual-side alignment: the learner removes weight quantization mismatch, while the rollout engine directly reduces the activation quantization error that cannot be fixed by training-side QDQ alone.

\section{Experiments}
\label{sec:experiments}

We evaluate the proposed dual-side alignment framework on full-parameter GRPO training of a large-scale MoE model under NVFP4 W4A4 rollout.
We first describe the experimental setup (Section~\ref{sec:exp-setup}), followed by the main RL training results (Section~\ref{sec:main-results}), ablation studies (Section~\ref{sec:ablation}), efficiency analysis (Section~\ref{sec:efficiency}), and activation mismatch visualization (Section~\ref{sec:visualization}).

\subsection{Experimental Setup}
\label{sec:exp-setup}

\paragraph{Model Configuration.}
We conduct experiments on MoE 25B-A2.8B model.
Training and rollout are performed on a cluster of NVIDIA Blackwell GPUs.
We adopt a disaggregated architecture that decouples training and inference onto separate node pools.
The training backend uses Megatron-LM with tensor-model parallelism $\mathrm{TP}{=}2$, expert-model parallelism $\mathrm{EP}{=}16$, context parallelism $\mathrm{CP}{=}1$, pipeline parallelism $\mathrm{PP}{=}1$, and sequence parallelism $\mathrm{SP}{=}1$, yielding a data-parallel degree of $\mathrm{DP}{=}1$.
The rollout engine is SGLang with \texttt{trtllm\_mha} attention backend, $\mathrm{TP}{=}2$, and $\mathrm{DP}{=}1$.
Weights are synchronized from the trainer to the rollout engine at each checkpoint refresh.

\paragraph{RL Configuration.}
We train with Group Relative Policy Optimization (GRPO)~\citep{shao2024deepseekmath} as follows:
Each prompt is sampled with $N{=}16$ responses; the global batch size is 256 and the mini-batch size is 128.
We use a single PPO epoch per update with a learning rate of $1\mathrm{e-}6$ and AdamW optimizer ($\beta_1{=}0.9$, $\beta_2{=}0.95$, $\epsilon{=}1\mathrm{e-}15$).
The maximum prompt length is 3{,}072 tokens and the maximum generation length is 32{,}768 tokens.
We adopt asymmetric clipping with positive clip ratio $0.27$ and negative clip ratio $0.2$, a dynamic PG loss strategy with cumulative clip range of 90, and sample-averaged loss.
Gradient checkpointing is enabled with MoE-layer recomputation.
Optimizer states and main weights are offloaded to CPU memory to fit the model within the GPU memory budget.
All experiments run for 260 training steps and evaluation every 10 steps.

\paragraph{Datasets and evaluation.}
Training data consists of a mixture of mathematical reasoning and code generation prompts:
(i)~a curated math dataset combining O4-Mini, verified prover data, and DeepMath with Chinese-language entries filtered out;
(ii)~a filtered code dataset spanning diverse programming tasks.
We evaluate on two held-out benchmarks:
(i)~\textbf{AIME 2024/25 + HMMT 2025}, a suite of competition-level mathematics problems;
(ii)~\textbf{LiveCodeBench} (LCB), a continuously updated code generation benchmark with Python-only submissions evaluated via a unified evaluation service.
All evaluation results are reported as pass@1 accuracy averaged over multiple copies.

\begin{table}[htbp]
    \centering
    \caption{Pass@1 accuracy (\%) on four benchmarks. 
    \textbf{BF16 RL} is the full-precision baseline; \textbf{FP8 RL} applies FP8 quantization on both sides; \textbf{NVFP4 + BF16 Train} uses NVFP4 rollout with BF16 training and no alignment; \textbf{QUADS} is our proposed dual-side alignment pipeline in NVFP4 RL.
    Our method achieves accuracy on par with BF16 across all benchmarks, while the naive NVFP4 RL suffers severe degradation.
    For each benchmark, the best result is shown in bold and the second-best result is underlined.}
    \label{tab:main-results}
    \begin{tabular}{lccccc}
    \toprule
    \textbf{Configuration} & \textbf{LiveCodeBench} & \textbf{HMMT 2025} & \textbf{AIME 2024} & \textbf{AIME 2025} & \textbf{Average} \\
    \midrule
    BF16 RL & 63.05 & \textbf{66.04} & \textbf{83.54} & \underline{80.00} & 73.15 \\
    FP8 RL & \textbf{65.27} & \underline{63.54} & 82.91 & 79.79 & 72.88 \\
    Naive NVFP4 RL & 24.25 & 48.33 & 66.67 & 66.25 & 51.37 \\
    QUADS (ours) & \underline{64.79} & 62.92 & \underline{83.33} & \textbf{80.42} & 72.86 \\
    \bottomrule
    \end{tabular}
\end{table}

\subsection{Main Experiment}
\label{sec:main-results}

We compare the following configurations, for each configuration, we train for 260 steps and report pass@1 accuracy on four held-out benchmarks: AIME 2024, AIME 2025, HMMT 2025, and LiveCodeBench.

\begin{itemize}[leftmargin=*,itemsep=2pt,topsep=2pt]
    \item \textbf{BF16 RL}: BF16 rollout and BF16 training. This is the standard full-precision baseline with no quantization on either side.
    \item \textbf{FP8 RL}: FP8 rollout and FP8 training. Both weights and activations are quantized to FP8 at inference time and the trainer applies FP8 training.
    \item \textbf{Naive NVFP4 RL}: NVFP4 W4A4 rollout and BF16 training. Both weights and activations are quantized to E2M1 at inference time while the trainer operates in BF16.
    \item \textbf{QUADS} (ours): The dual-side alignment pipeline combining Asymmetric QAT (Section~\ref{sec:methods:w4a16-qat}) with inference-side residual activation compensation (Section~\ref{sec:methods:residual}).
\end{itemize}

\begin{figure}[htbp]
    \centering
    \begin{minipage}{0.48\linewidth}
        \centering
        \includegraphics[width=\linewidth]{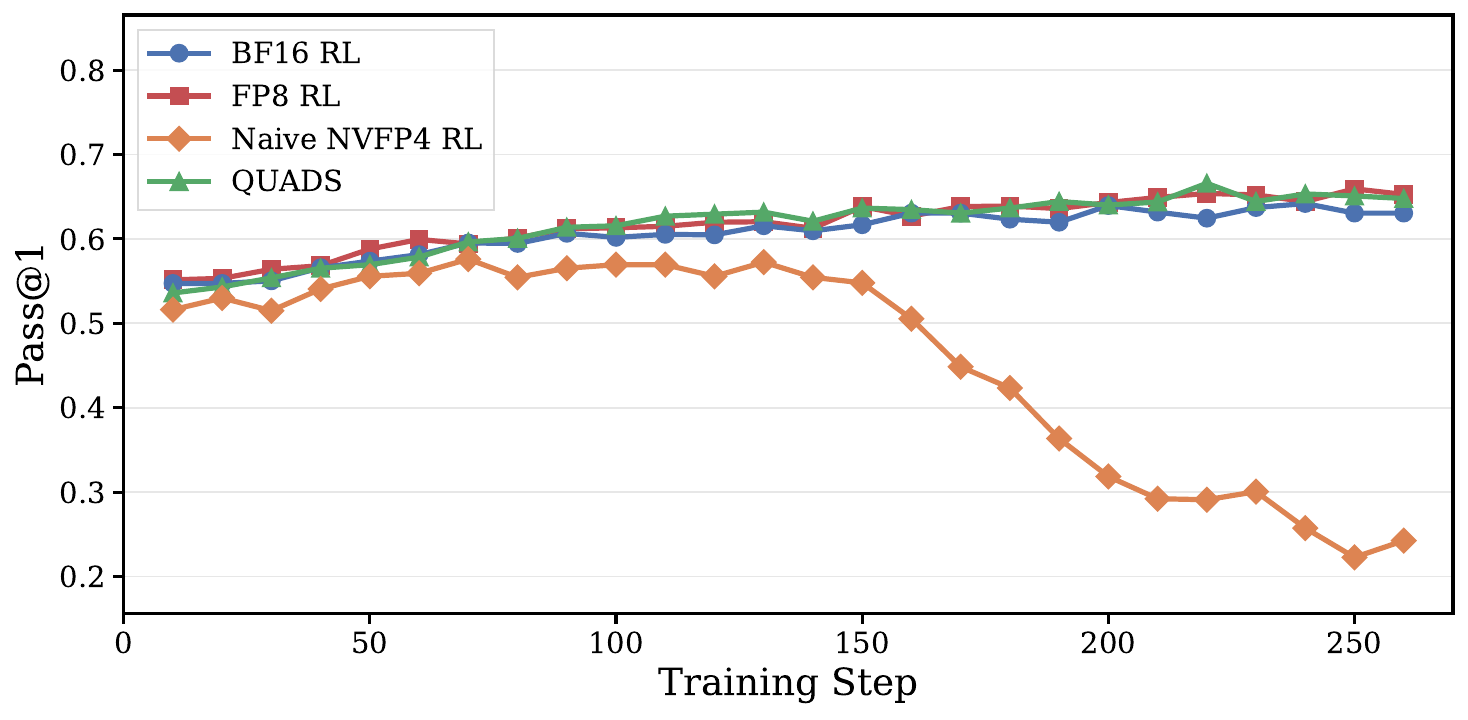}
        \centerline{\small (a) LiveCodeBench}
    \end{minipage}\hfill
    \begin{minipage}{0.48\linewidth}
        \centering
        \includegraphics[width=\linewidth]{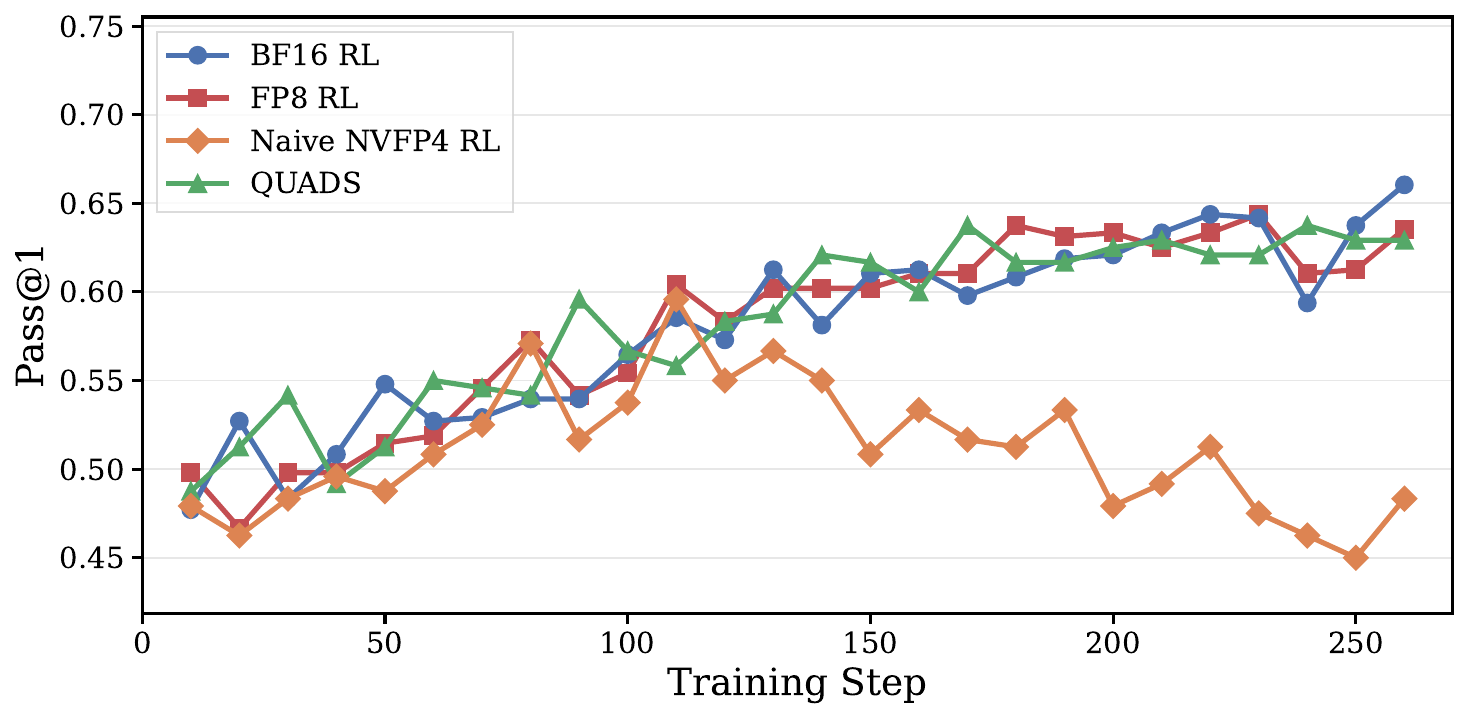}
        \centerline{\small (b) HMMT 2025}
    \end{minipage}
    \vspace{2mm}
    \begin{minipage}{0.48\linewidth}
        \centering
        \includegraphics[width=\linewidth]{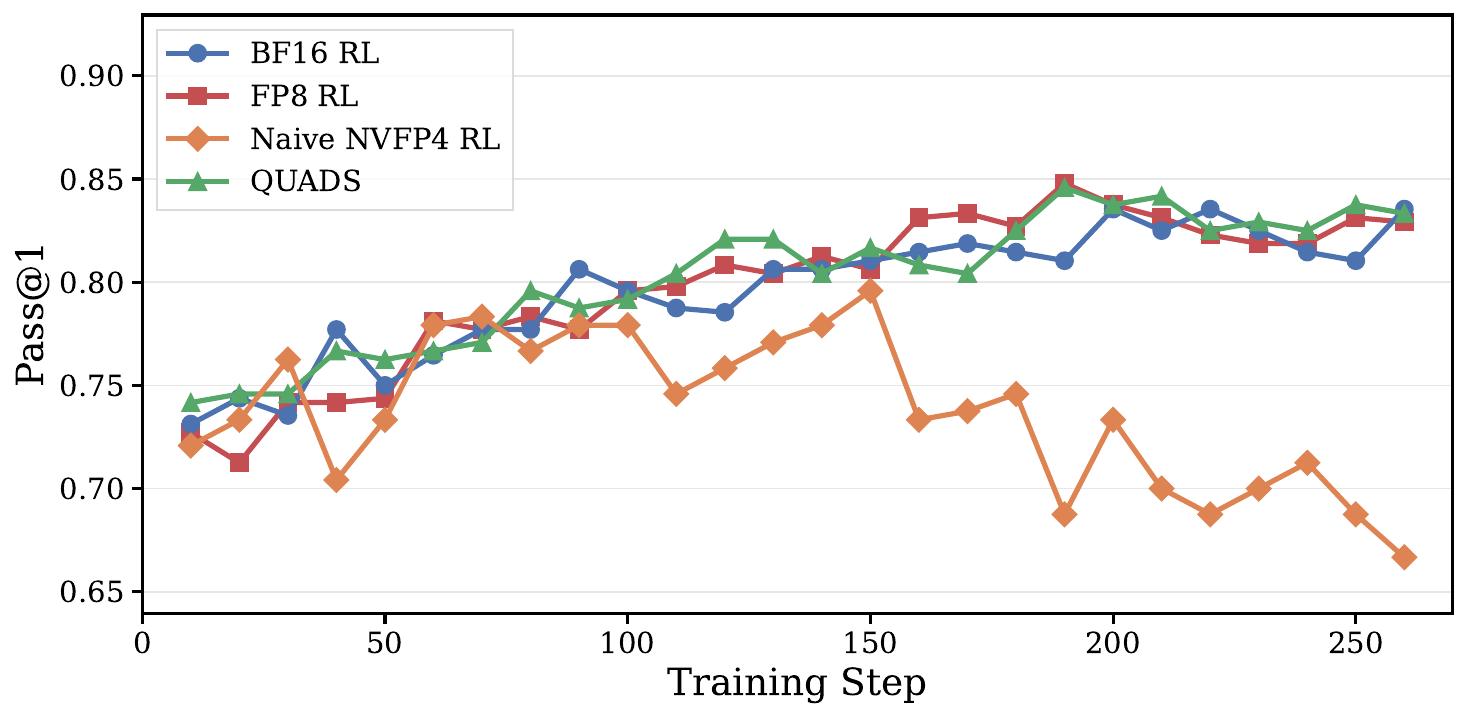}
        \centerline{\small (c) AIME 2024}
    \end{minipage}\hfill
    \begin{minipage}{0.48\linewidth}
        \centering
        \includegraphics[width=\linewidth]{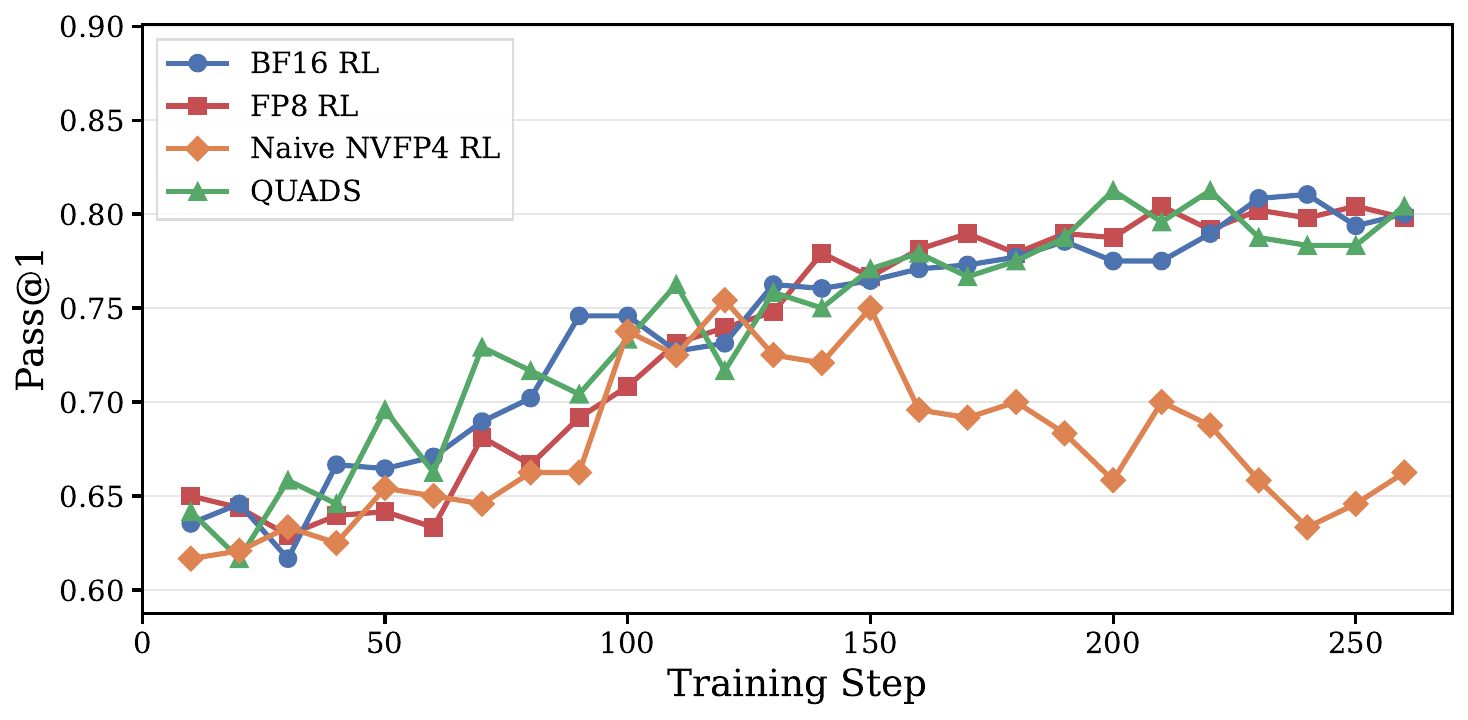}
        \centerline{\small (d) AIME 2025}
    \end{minipage}
    \caption{Pass@1 score curves on four held-out benchmarks.
    \textbf{BF16} (blue) and \textbf{FP8 RL} (red) provide strong baselines.
    \textbf{QUADS} (green, ours) tracks both baselines closely on every benchmark.
    \textbf{NVFP4 + BF16 Train} (orange, naive W4A4) has lower initial point and collapses after roughly 150 steps.}
    \label{fig:main-results}
\end{figure}

\paragraph{Key observations.} As illustrated in Figure~\ref{fig:main-results} and Table~\ref{tab:main-results}, we summarize the following:
\begin{itemize}[leftmargin=*,itemsep=2pt,topsep=2pt]
    \item \textbf{BF16} serves as the upper-bound reference, exhibiting steady score growth across all four benchmarks.
    \item \textbf{FP8 RL} is the standard low-precision baseline; with FP8 training and FP8 rollout, its learning curve closely tracks BF16 RL.
    \item \textbf{NVFP4 + BF16 Train} (no alignment) collapses after roughly 150 steps on every benchmark due to unmitigated training--inference mismatch (Figure~\ref{fig:main-results}, orange curves), confirming that a naive NVFP4 setup is incompatible with stable RL training.
    \item \textbf{QUADS} (ours) recovers accuracy to within the variance of the BF16 baseline across all four benchmarks, demonstrating that training-side weight alignment combined with inference-side residual activation compensation is both necessary and sufficient for stable NVFP4 RL.
\end{itemize}

A central failure mode of naive NVFP4 RL is the growing discrepancy between rollout and trainer log-probabilities, which directly biases importance ratios in GRPO.
We therefore track the maximum per-step log-probability difference $\|\log \pi_{\theta}^{\mathrm{infer}} - \log \pi_{\theta}^{\mathrm{train}}\|_{\max}$ across three configurations: \textbf{Naive NVFP4 RL} (no alignment), \textbf{QUADS} (dual-side alignment), and \textbf{BF16 RL} (full-precision reference).
Figure~\ref{fig:logprob-gap} reports this metric over 260 training steps.
Naive NVFP4 RL maintains the largest gap ($\sim$1.3) throughout training, explaining the policy-gradient instability observed in Figure~\ref{fig:main-results}.
QUADS consistently lowers the gap to $\sim$0.86, substantially closing the train--inference discrepancy.
The BF16 baseline ($\sim$0.74) represents the irreducible engine-drift floor between Megatron-LM and SGLang, confirming that our dual-side alignment approaches this lower bound.

\subsection{Ablation Studies}
\label{sec:ablation}

\subsubsection{W4A4 QAT vs.\ W4A16 QAT}
\label{sec:ablation-qat}

\begin{figure}[htbp]
    \centering
    \begin{minipage}{0.48\linewidth}
        \centering
        \includegraphics[width=\linewidth]{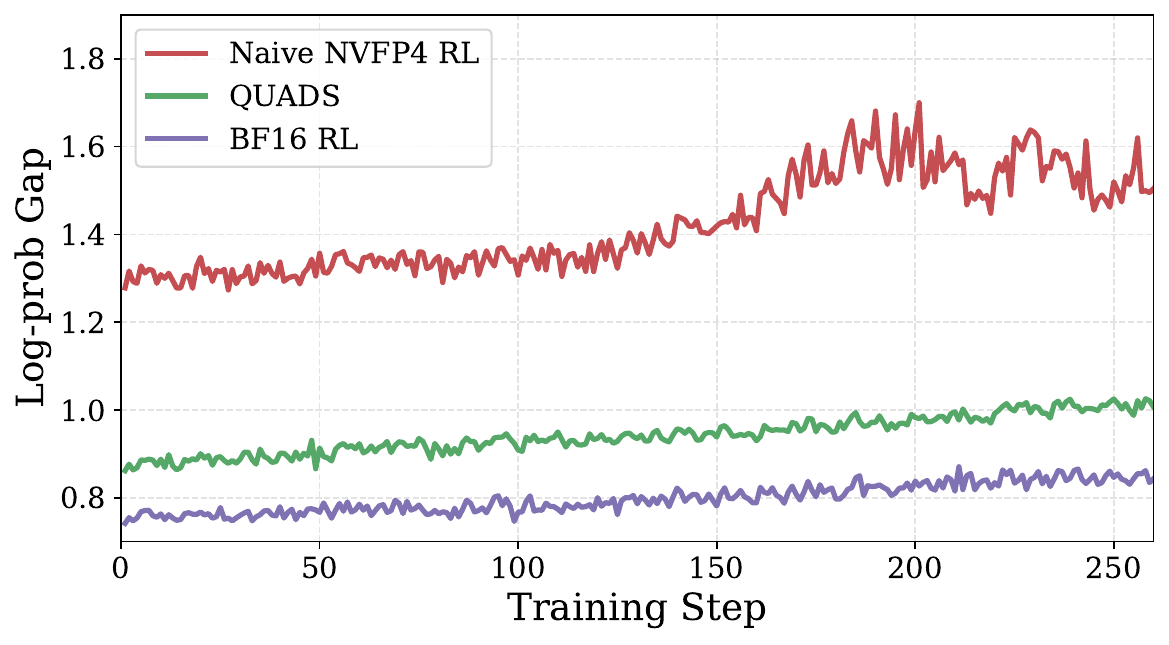}
        \captionof{figure}{Maximum log-probability gap $\|\log \pi_{\theta}^{\mathrm{infer}} - \log \pi_{\theta}^{\mathrm{train}}\|_{\max}$ over training steps.
        \textbf{Naive NVFP4 RL} (red) sustains the largest gap; \textbf{QUADS} (green, ours) substantially reduces it toward the \textbf{BF16 RL} (purple) engine-drift floor.}
        \label{fig:logprob-gap}
    \end{minipage}\hfill
    \begin{minipage}{0.48\linewidth}
        \centering
        \includegraphics[width=\linewidth]{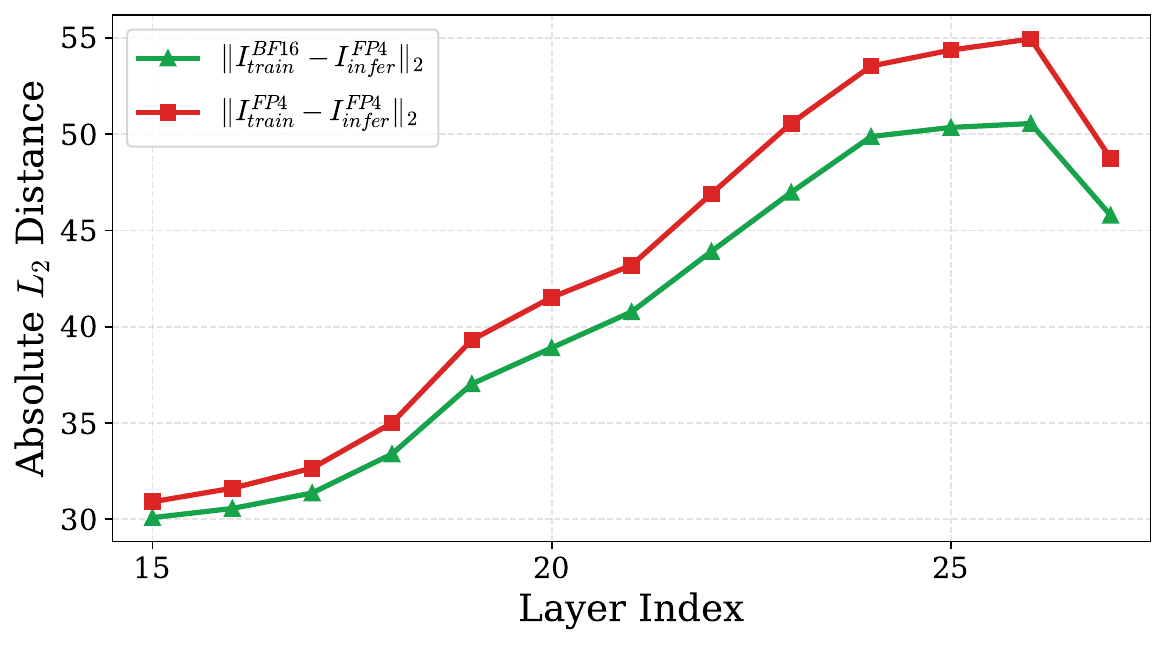}
        \captionof{figure}{Layer-wise MoE-input mismatch under two QAT configurations on matched requests. \textbf{Green}: Asymmetric W4A16 QAT. \textbf{Red}: Symmetric W4A4 QAT. Symmetric W4A4 QAT shows larger mismatch.}
        \label{fig:w4a16-vs-w4a4-qat}
    \end{minipage}
\end{figure}

We ablate symmetric W4A4 QAT, which fake-quantizes both weights and activations during training, against asymmetric W4A16 QAT, which fake-quantizes weights only, on matched requests.
Figure~\ref{fig:w4a16-vs-w4a4-qat} reports layer-wise MoE-input mismatch under two training--inference configurations:
\begin{itemize}[leftmargin=*,itemsep=2pt,topsep=2pt]
    \item \textbf{Asymmetric W4A16 QAT} (green): $\|\mathbf{I}_{\mathrm{train}}^{\mathrm{BF16}} - \mathbf{I}_{\mathrm{infer}}^{\mathrm{FP4}}\|_2$, with weight-aligned training and W4A4 inference.
    \item \textbf{Symmetric W4A4 QAT} (red): $\|\mathbf{I}_{\mathrm{train}}^{\mathrm{FP4}} - \mathbf{I}_{\mathrm{infer}}^{\mathrm{FP4}}\|_2$, with both operands fake-quantized during training and W4A4 inference.
\end{itemize}
Symmetric W4A4 QAT (red) exhibits the largest layer-wise gap, confirming the analysis in Section~\ref{sec:methods:w4a16-qat}: coarse E2M1 activation fake-quantization amplifies engine drift rather than reducing it.
By contrast, asymmetric W4A16 QAT (green) reduces the activation error on each layer, verifying our asymmetric training design.

\subsubsection{Effect of Dual-side Alignment }
\label{sec:ablation-residual}

In this section, we evaluate the effect of our dual-side alignment method by comparing three configurations:

\begin{itemize}[leftmargin=*,itemsep=2pt,topsep=2pt]
    \item \textbf{Naive NVFP4}: NVFP4 rollout and BF16 training, without alignment on either side.
    \item \textbf{W4A16 QAT Only}: NVFP4 rollout and W4A16 QAT, only with training-side alignment.
    \item \textbf{QUADS} training-side weight alignment \emph{and} inference-side residual compensation, with both components.
\end{itemize}

\begin{figure}[htbp]
    \centering
    \begin{minipage}{0.48\linewidth}
        \centering
        \includegraphics[width=\linewidth]{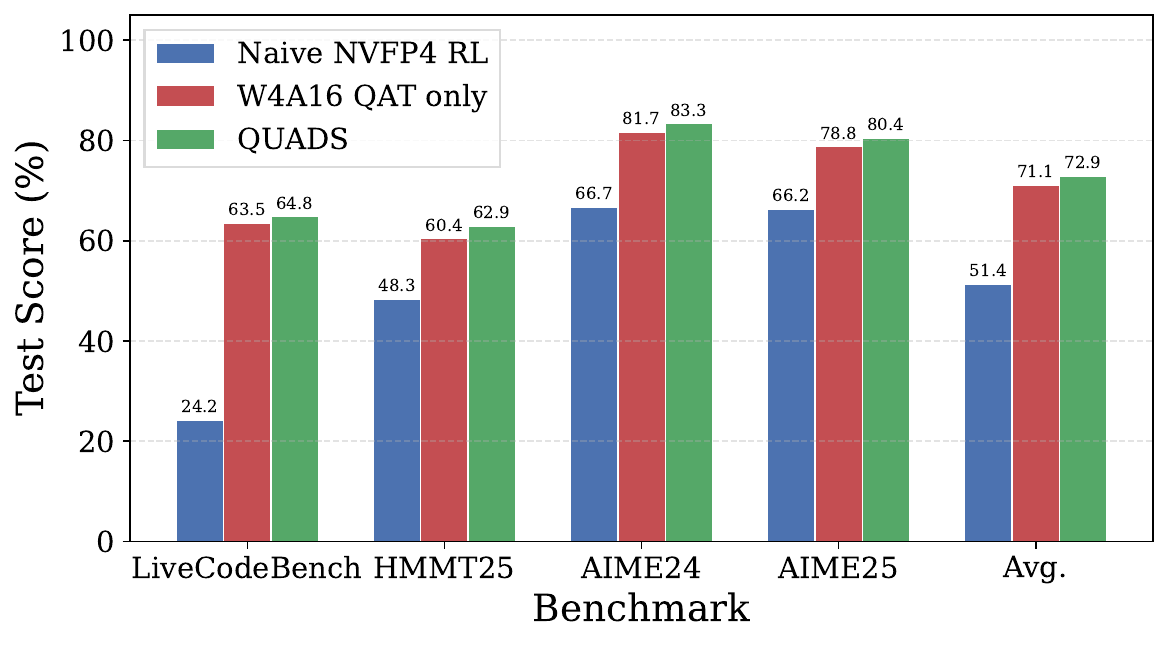}
        \centerline{\small (a) Benchmark Scores}
    \end{minipage}\hfill
    \begin{minipage}{0.48\linewidth}
        \centering
        \includegraphics[width=\linewidth]{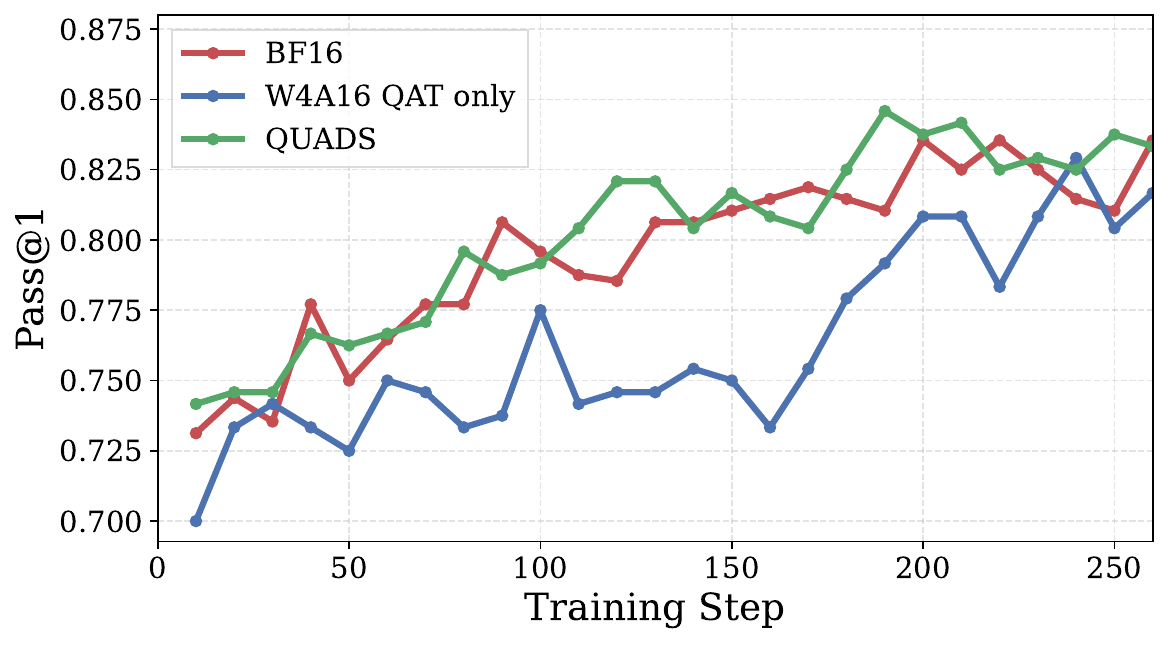}
        \centerline{\small (b) AIME24 Test Score}
    \end{minipage}
    \caption{Ablation of training-side W4A16 QAT and inference-side residual compensation.
    \textbf{(a)}~Final pass@1 scores across held-out benchmarks: naive NVFP4 RL without alignment degrades severely (51.4\% average), W4A16 QAT alone recovers most of the gap (71.1\%), and adding residual compensation yields further consistent gains on every benchmark (72.9\%).
    \textbf{(b)}~AIME24 score over training steps: W4A16 QAT only (red) grows more slowly and remains below the BF16 baseline (blue), whereas W4A16 QAT + Residual (green) tracks BF16 closely throughout training.}
    \label{fig:ablation-residual}
\end{figure}

As shown in Figure~\ref{fig:ablation-residual}, each alignment component contributes an independent gain.
Figure~\ref{fig:ablation-residual}(a) shows that naive NVFP4 + BF16 Train without any alignment suffers severe accuracy degradation across all benchmarks, confirming the failure mode analyzed in Section~\ref{sec:motivating-exp}.
W4A16 QAT alone stabilizes training and recovers most of the accuracy, yet still leaves a noticeable gap to the BF16 baseline.
Adding inference-side residual compensation on top of W4A16 QAT further closes this gap: the corrected activation tensors more closely approximate the BF16 reference, reducing $\delta_t^{\mathrm{act}}$ and improving every benchmark score.
Figure~\ref{fig:ablation-residual}(b) corroborates this at the trajectory level on AIME24---W4A16 QAT only lags BF16 throughout training, while the full pipeline restores evaluation growth to near-baseline levels.
These results confirm that both components contribute independently and complementarily to the final accuracy.

\subsubsection{Channel Selection Ratio}
\label{sec:ablation-channel}

The residual compensation module selects the top-$k$\% of channels by residual magnitude for correction (Section~\ref{sec:methods:residual}).
We study the sensitivity to $k$ by varying the selection ratio and measuring the maximum post-correction log-probability difference $\|\log \pi_{\theta}^{\mathrm{infer}} - \log \pi_{\theta}^{\mathrm{train}}\|_{\max}$ between the training and inference engines.

\begin{table}[htbp]
\centering
\caption{Maximum log-probability difference between training and inference engines under different residual channel selection ratios.}
\label{tab:channel-ratio}
\begin{tabular}{lcccccc}
\toprule
\textbf{Residual Channel Ratio} & Naive NVFP4 RL  & 12.5\% & 25\% & 50\% & 75\% & 100\% \\
\midrule
$\|\log \pi_{\theta}^{\mathrm{infer}} - \log \pi_{\theta}^{\mathrm{train}}\|_{\max}$ & 1.273  & 0.923 & 0.901 & 0.854 & 0.812 & 0.759\\
\bottomrule
\end{tabular}
\end{table}

Table~\ref{tab:channel-ratio} reports the maximum log-probability difference across training steps for each selection ratio.
As the residual channel ratio increases from 12.5\% to 75\%, the logprob gap decreases, indicating that correcting more channels further aligns the inference engine with the training engine.
However, larger ratios also incur proportionally higher compute overhead, so the optimal $k$ balances accuracy recovery against throughput cost.
\subsection{Efficiency Analysis}
\label{sec:efficiency}

\subsubsection{Residual Compensation FLOPs Cost} 
\label{sec:efficiency-flops}

The residual compensation module introduces additional computation on top of the standard NVFP4 W4A4 forward pass.
For each transformer layer, the correction involves: (i) extracting the top-$k$\% residual channels from the activation tensor, (ii) re-quantizing the residual to E2M1, and (iii) adding the corrected residual back to the quantized output.
The additional FLOPs are dominated by the element-wise operations on the selected channels, which scale as $\mathcal{O}(k \cdot d \cdot s)$ per layer, where $d$ is the hidden dimension and $s$ is the sequence length.
With $k{=}50\%$, the residual module adds less than 10\% FLOPs overhead relative to the base W4A4 GEMM, making it computationally lightweight.

\subsubsection{Rollout Throughput Comparison}
\label{sec:efficiency-throughput}

We measure decode throughput (tokens/s) on the SGLang engine for three rollout configurations:
\begin{itemize}[leftmargin=*,itemsep=2pt,topsep=2pt]
    \item \textbf{FP8 W8A8}: FP8 rollout as the current production baseline.
    \item \textbf{NVFP4 W4A4}: Plain NVFP4 rollout without residual compensation.
    \item \textbf{NVFP4 W4A4 + Residual} (ours): NVFP4 rollout with inference-side activation error compensation.
\end{itemize}

\begin{figure}[htbp]
    \centering
    \begin{subfigure}[t]{0.48\linewidth}
        \centering
        \includegraphics[width=\linewidth]{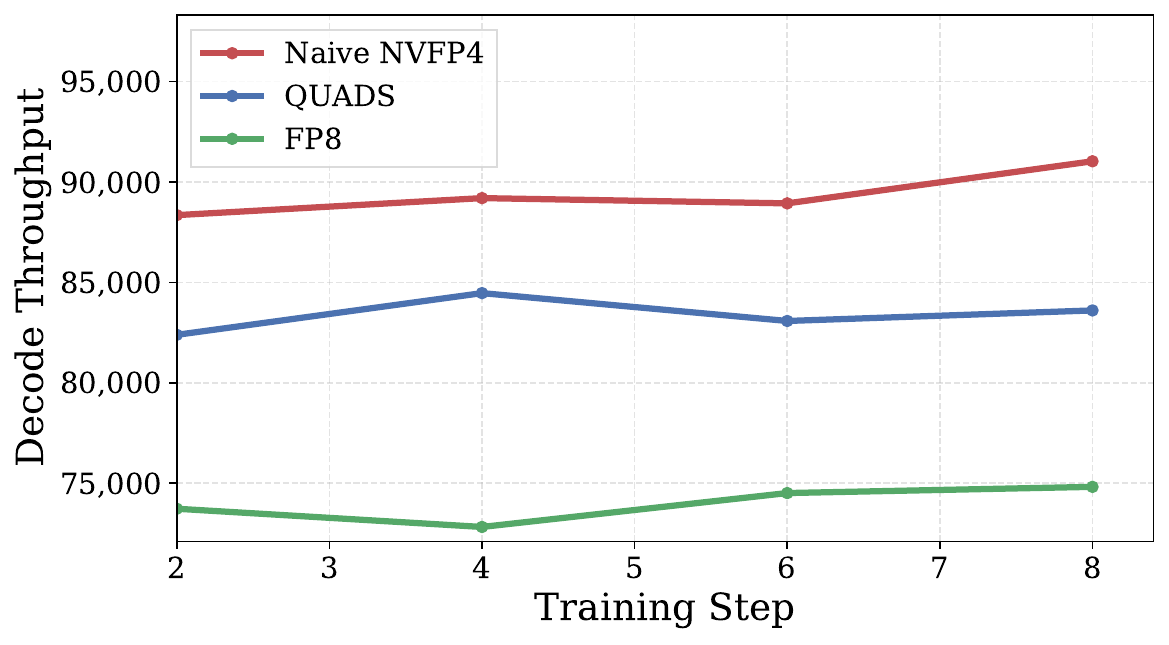}
        \caption{Decode throughput on SGLang across RL training steps.
        \textbf{Naive NVFP4 W4A4} (red) consistently achieves the highest throughput.
        \textbf{QUADS} (blue, ours) incurs a modest $\sim$5--9\% slowdown relative to plain W4A4 but remains substantially faster than \textbf{FP8} (green).}
        \label{fig:rollout-throughput}
    \end{subfigure}
    \hfill
    \begin{subfigure}[t]{0.48\linewidth}
        \centering
        \includegraphics[width=\linewidth]{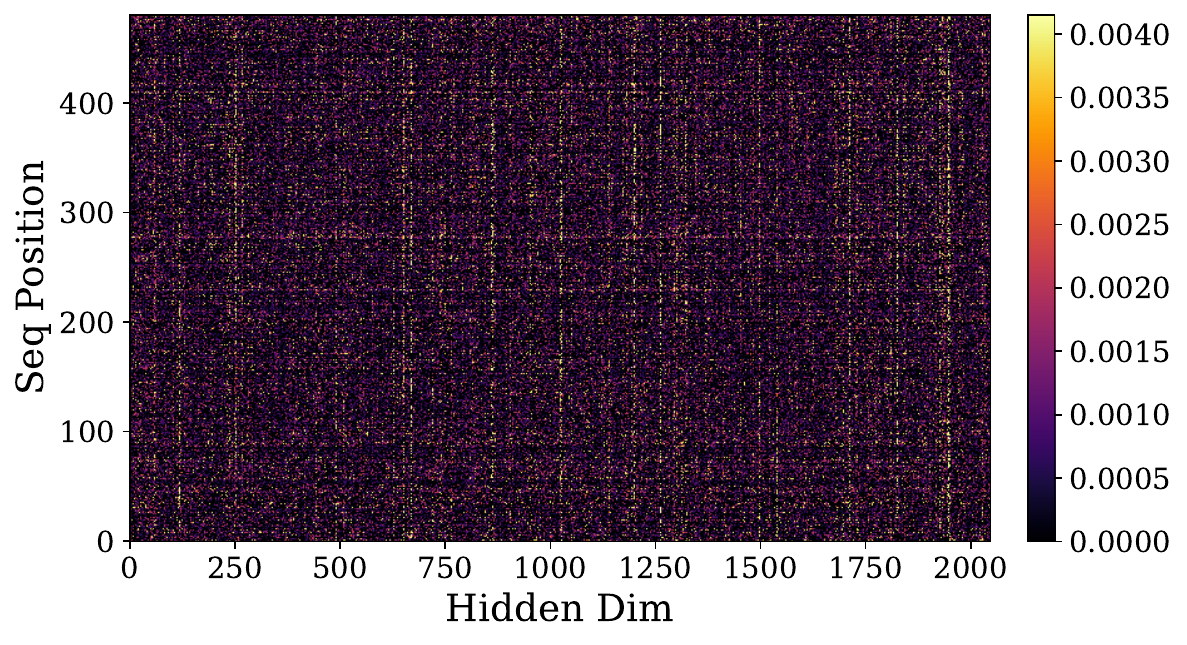}
        \caption{Element-wise activation mismatch $|\mathbf{A}_{\mathrm{train}}^{\mathrm{BF16}} - \mathbf{A}_{\mathrm{infer}}^{\mathrm{BF16}}|$ at Layer~1.
        Even without any quantization, structured non-zero residuals are clearly visible, revealing that kernel-level implementation differences introduce an irreducible \emph{engine drift}.}
        \label{fig:train-infer-mismatch}
    \end{subfigure}
    \caption{(a) Rollout throughput comparison across training steps. (b) BF16 Train--infer activation mismatch heatmap at Layer~27, showing irreducible engine drift between Megatron-LM and SGLang even in BF16.}
    \label{fig:throughput-and-mismatch}
\end{figure}

Figure~\ref{fig:rollout-throughput} reports throughput at several steps during RL training.
Plain NVFP4 W4A4 rollout delivers roughly $1.3\times$ the FP8 throughput, reflecting the benefit of FP4 Tensor Core kernels under our production workload.
Adding residual compensation reduces throughput by only $\sim$5--9\% compared with plain W4A4, owing to fused Triton kernels that combine quantization, residual extraction, and masked reconstruction with minimal launch overhead.
The full pipeline therefore retains a clear speed advantage over FP8 throughput by $\sim$16\%, indicating that the accuracy gains from residual compensation come at an acceptable throughput cost.

\subsection{Visualization}
\label{sec:visualization}

\subsubsection{Training--Inference Activation Mismatch}
\label{sec:vis-mismatch}

To understand the fundamental source of train-infer mismatch, we visualize the element-wise activation differences between the BF16 training engine (Megatron-LM) and the BF16 inference engine (SGLang) on identical input batches.
Figure~\ref{fig:train-infer-mismatch} plots $|\mathbf{A}_{\mathrm{train}}^{\mathrm{BF16}} - \mathbf{A}_{\mathrm{infer}}^{\mathrm{BF16}}|$ as a heatmap at Layer~1, where each pixel corresponds to a single element indexed by sequence position and hidden dimension.
Even without any quantization, the two engines produce non-identical activations due to differences in kernel implementations, accumulation order, and numerical rounding.
The heatmap reveals that the mismatch is not uniformly distributed: certain channels exhibit consistently larger residuals, suggesting that specific hidden dimensions are more sensitive to implementation-level numerical discrepancies.
This irreducible \emph{engine drift} constitutes the lower bound of training--inference mismatch: any quantization scheme can at best match this gap, but never eliminate it.
Our dual-side alignment framework is designed to approach this lower bound by minimizing the additional mismatch introduced by quantization on both sides.

\section{Related Work}
\label{sec:related-work}

\subsection{Reinforcement Learning for Large Language Models}

Reinforcement learning from human feedback (RLHF) and its verifiable-reward variant (RLVR) have become standard post-training stages for aligning large language models~\citep{schulman2017ppo,shao2024deepseekmath}.
PPO~\citep{schulman2017ppo} introduced the clipped surrogate objective that bounds policy updates within a trust region, while GRPO~\citep{shao2024deepseekmath} removes the value model and normalizes advantages within a sampled response group, substantially simplifying the training pipeline for reasoning tasks.
As chain-of-thought traces grow longer, autoregressive rollout on a separate inference engine dominates end-to-end wall-clock time, motivating a line of work on accelerating the RL loop through system-level optimizations~\citep{yao2025offpolicy,li2026qurl,xi2026jetrl}.

\subsection{Low-Precision Inference and RL Rollout}

Low-precision inference has matured from INT8~\citep{li2026qurl} through FP8~\citep{deepgemm2025,xi2026jetrl,qiu2026fp8rl} to the emerging FP4 tier.
FP8 E4M3, natively supported on Hopper and Blackwell Tensor Cores, provides 256 representable levels per sign and has become the default precision for high-throughput LLM serving.
NVFP4~\citep{alvarez2025nvfp4blog,nvidia2025nvfp4pretrain} further halves the bit width to E2M1 with block-scaled E4M3 scaling, targeting $\sim$4$\times$ GEMM throughput over BF16 on Blackwell FP4 Tensor Cores~\citep{jarmusch2025blackwell}.
When such quantized inference is used for RL rollout, quantization-aware training (QAT) can reduce rollout--trainer mismatch by inserting fake quantization--dequantization nodes into the learner~\citep{gu2026qarl}.
For weight-only quantization, QAT is relatively straightforward: both engines multiply by the same QDQ-aligned weight, and the straight-through estimator propagates gradients through the rounding operator.
QeRL~\citep{huang2025qerl} applies NVFP4 weight-only quantization to RL but keeps activations in BF16, executing FP16 GEMMs rather than FP4 Tensor Core operations---thus sidestepping activation quantization entirely.
Symmetric W4A4 QAT, which fake-quantizes both weights and activations, appears to offer full alignment but in practice amplifies the engine drift $\boldsymbol{\eta}$ through the coarse E2M1 decision boundaries (Section~\ref{sec:methods:w4a16-qat}).
Our work identifies this operand asymmetry and proposes asymmetric W4A16 QAT combined with inference-side residual compensation, enabling the stable full-parameter MoE RL with native W4A4 FP4 GEMM on Blackwell hardware.

\subsection{Training--Inference Mismatch in RL}

When rollout and training run on separate engines at different precisions, the per-token log-probability gap $\delta_t$ corrupts importance ratios and destabilizes policy-gradient updates~\citep{qi2025precisionrl,yao2025offpolicy}.
A series of recent works address this mismatch at the FP8 and INT8 precision tiers.
QuRL~\citep{li2026qurl} quantizes the actor for generation while keeping full-precision gradient updates, using adaptive clipping to control importance ratios.
Jet-RL~\citep{xi2026jetrl} enforces a unified FP8 precision flow across training and rollout, eliminating the precision gap at the cost of quantizing the learner.
FP8-RL~\citep{qiu2026fp8rl} provides a practical low-precision stack with truncated importance sampling, while QaRL~\citep{gu2026qarl} aligns rollout and training through rollout-aware quantization-aware training.
AIS~\citep{zhou2026ais} introduces adaptive importance sampling that dynamically adjusts clipping thresholds based on observed ratio distributions.
All of these methods operate on the comparatively fine FP8 or INT8 grid, where $|\delta_t|$ remains tractable; none targets native W4A4 FP4 Tensor Core inference, where the E2M1 grid is $32\times$ coarser than E4M3 and importance-sampling corrections alone prove insufficient.

\section{Conclusion}
\label{sec:conclusion}

We study NVFP4 rollout for Mixture-of-Experts (MoE) Reinforcement Learning on NVIDIA Blackwell.
Although native W4A4 FP4 GEMMs offer higher rollout throughput than FP8, directly combining NVFP4 rollout with BF16 training causes unstable GRPO training.
Through training--inference error analysis and controlled ablations, we identify activation quantization, rather than weight quantization, as the dominant source of FP4 RL instability.

Based on this finding, we propose dual-side alignment: asymmetric W4A16 quantization-aware training aligns the trainer-side weight path, while residual activation compensation reduces rollout-side activation error without leaving native W4A4 GEMMs.
On full-parameter GRPO for the MoE model, the proposed NVFP4 pipeline reaches BF16-level accuracy, improves average pass@1 by 21.49 percentage points over naive NVFP4 RL, and retains a 16\% rollout throughput advantage over FP8.
Future work includes more adaptive residual-channel selection and extending this alignment principle to other RL algorithms and model architectures.

\bibliographystyle{plainnat}
\bibliography{reference}

@misc{alvarez2025nvfp4blog,
  title         = {Introducing {NVFP4} for Efficient and Accurate Low-Precision Inference},
  author        = {Alvarez, Eduardo and Almog, Omri and Chung, Eric and Layton, Simon and Stosic, Dusan and Krashinsky, Ronny and Aubrey, Kyle},
  year          = {2025},
  howpublished  = {NVIDIA Technical Blog},
  url           = {https://developer.nvidia.com/blog/introducing-nvfp4-for-efficient-and-accurate-low-precision-inference/}
}

@article{nvidia2025nvfp4pretrain,
  title   = {Pretraining Large Language Models with {NVFP4}},
  author  = {{NVIDIA}},
  journal = {arXiv preprint arXiv:2509.25149},
  year    = {2025}
}

@article{jarmusch2025blackwell,
  title   = {Microbenchmarking {NVIDIA}'s Blackwell Architecture: An In-Depth Architectural Analysis},
  author  = {Jarmusch, Aaron and Chandrasekaran, Sunita},
  journal = {arXiv preprint arXiv:2512.02189},
  year    = {2025}
}

@inproceedings{schulman2017ppo,
  title     = {Proximal Policy Optimization Algorithms},
  author    = {Schulman, John and Wolski, Filip and Dhariwal, Prafulla and Radford, Alec and Klimov, Oleg},
  booktitle = {arXiv preprint arXiv:1707.06347},
  year      = {2017}
}

@article{shao2024deepseekmath,
  title   = {{DeepSeekMath}: Pushing the Limits of Mathematical Reasoning in Open Language Models},
  author  = {Shao, Zhihong and Wang, Peiyi and Zhu, Qihao and Xu, Runxin and Song, Junxiao and Bi, Xiao and Zhang, Haowei and Ming, Y. K. and Zhang, Y. Wu and Guo, Daya and others},
  journal = {arXiv preprint arXiv:2402.03300},
  year    = {2024}
}

@article{qi2025precisionrl,
  title   = {Defeating the Training-Inference Mismatch via {FP16}},
  author  = {Qi, Penghui and Liu, Zichen and Zhou, Xiangxin and Pang, Tianyu and Du, Chao and Lee, Wee Sun and Lin, Min},
  journal = {arXiv preprint arXiv:2510.26788},
  year    = {2025}
}

@misc{yao2025offpolicy,
  title        = {Your Efficient {RL} Framework Secretly Brings You Off-Policy {RL} Training},
  author       = {Yao, Feng and Liu, Liyuan and Zhang, Dinghuai and Dong, Chengyu and Shang, Jingbo and Gao, Jianfeng},
  year         = {2025},
  howpublished = {Notion Blog},
  url          = {https://fengyao.notion.site/off-policy-rl}
}

@article{huang2025qerl,
  title   = {{QeRL}: Beyond Efficiency--Quantization-enhanced Reinforcement Learning for {LLMs}},
  author  = {Huang, Wei and Ge, Yi and Yang, Shuai and Xiao, Yicheng and Mao, Huizi and Lin, Yujun and Ye, Hanrong and Liu, Sifei and Cheung, Ka Chun and Yin, Hongxu and Lu, Yao and Qi, Xiaojuan and Han, Song and Chen, Yukang},
  journal = {arXiv preprint arXiv:2510.11696},
  year    = {2025}
}

@article{li2026qurl,
  title   = {{QuRL}: Efficient Reinforcement Learning with Quantized Rollout},
  author  = {Li, Yuhang and Elangovan, Reena and Dong, Xin and Panda, Priyadarshini and Khailany, Brucek},
  journal = {arXiv preprint arXiv:2602.13953},
  year    = {2026}
}

@article{gu2026qarl,
  title   = {{QaRL}: Rollout-Aligned Quantization-Aware {RL} for Fast and Stable Training under Training--Inference Mismatch},
  author  = {Gu, Hao and Wang, Hao and Liu, Jiacheng and Li, Lujun and Zhu, Qiyuan and Liu, Bei and Xu, Binxing and Wang, Lei and Yang, Xintong and Lin, Sida and Han, Sirui and Guo, Yike},
  journal = {arXiv preprint arXiv:2604.07853},
  year    = {2026}
}

@article{xi2026jetrl,
  title   = {{Jet-RL}: Enabling On-Policy {FP8} Reinforcement Learning with Unified Training and Rollout Precision Flow},
  author  = {Xi, Haocheng and Ruan, Charlie and Liao, Peiyuan and Lin, Yujun and Cai, Han and Zhao, Yilong and Yang, Shuo and Keutzer, Kurt and Han, Song and Zhu, Ligeng},
  journal = {arXiv preprint arXiv:2601.14243},
  year    = {2026}
}

@article{qiu2026fp8rl,
  title   = {{FP8-RL}: A Practical and Stable Low-Precision Stack for {LLM} Reinforcement Learning},
  author  = {Qiu, Zhaopeng and Yu, Shuang and Zhang, Jingqi and Zhang, Shuai and Huang, Xue and Yang, Jingyi and Lai, Junjie},
  journal = {arXiv preprint arXiv:2601.18150},
  year    = {2026}
}

@article{zhou2026ais,
  title   = {{AIS}: Adaptive Importance Sampling for Quantized {RL}},
  author  = {Zhou, Jiajun and Shao, Wei and Zheng, Lingchao and Fan, Yuwei and Wong, Ngai},
  journal = {arXiv preprint arXiv:2605.13907},
  year    = {2026}
}

@misc{deepgemm2025,
  title        = {{DeepGEMM}: Clean and Efficient {FP8} {GEMM} Kernels with Fine-Grained Scaling},
  author       = {Zhao, Chenggang and Xu, Zhean and Zhao, Liang and Li, Jiashi and Xu, Chenhao and Xu, Anyi and Liu, Shengyu and Zhou, Kexing and Yu, Kuai},
  year         = {2025},
  howpublished = {GitHub},
  url          = {https://github.com/deepseek-ai/DeepGEMM}
}

\end{document}